\documentclass[10pt,twocolumn,letterpaper]{article}

\usepackage{cvpr}              
\usepackage[dvipsnames]{xcolor}
\usepackage{float}

\definecolor{cvprblue}{rgb}{0.21,0.49,0.74}
\usepackage[pagebackref,breaklinks,colorlinks,citecolor=cvprblue]{hyperref}

\usepackage{array}
\newcolumntype{R}[1]{>{\hbox to #1\bgroup\hfill$}c<{$\egroup}}

\def\paperID{15904} 
\def\confName{CVPR}
\def\confYear{2024}

\title{The Devil is in the Details: StyleFeatureEditor for Detail-Rich StyleGAN Inversion and High Quality Image Editing}

\author{
Denis Bobkov\textsuperscript{\rm 1} \quad Vadim Titov\textsuperscript{\rm 2} \quad Aibek Alanov\textsuperscript{\rm 1,2} \quad Dmitry Vetrov\textsuperscript{\rm 3}\\
	\small \textsuperscript{\rm 1}HSE University \quad \textsuperscript{\rm 2}AIRI \quad \textsuperscript{\rm 3}Constructor University, Bremen\\
	{\small \tt dnbobkov@edu.hse.ru, titov@2a2i.org, aalanov@hse.ru, dvetrov@constructor.university} 
}

\usepackage[accsupp]{axessibility}
\usepackage{xr}
\begin{document}
\twocolumn[{
\renewcommand\twocolumn[1][]{#1}
\vspace{-1.5cm}
\maketitle
\begin{center}
    \centering
    \vspace{-0.25cm}
    \includegraphics[width=\textwidth]{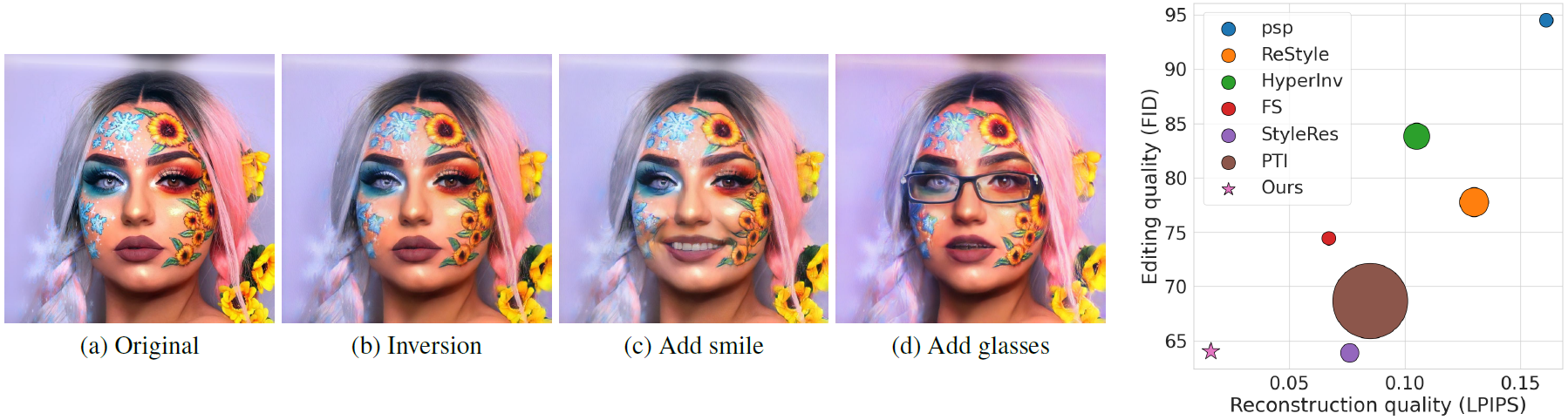}
    \vspace{-0.7cm}
    \captionof{figure}{{\bf Editing examples and graphical 
 comparison for StyleFeatureEditor.} {\it Our approach takes a real image, inverts it to the StyleGAN latent space, edits the found latents, and synthesises the edited image. On the left, we present examples of our approach, while on the right, we display a comparison with previous approaches. To evaluate inversion quality, we used LPIPS$\downarrow$. Additionally, to compare the editing capabilities, we compute FID$\downarrow \hspace{0.5mm}$  for 3 editing directions (see \ref{quant}) and average them with coefficients equal to the average FID per editing direction. The size of markers indicates the inference time of the method, with larger markers indicating a higher time. StyleFeatureEditor capable of reconstructing even finer image details and preserving them during editing.}}
\label{fig:teaser}
\end{center}
}] 

\begin{abstract}
 \vspace{-1.5cm}
The task of manipulating real image attributes through StyleGAN inversion has been extensively researched. This process involves searching latent variables from a well-trained StyleGAN generator that can synthesize a real image, modifying these latent variables, and then synthesizing an image with the desired edits. A balance must be struck between the quality of the reconstruction and the ability to edit. Earlier studies utilized the low-dimensional W-space for latent search, which facilitated effective editing but struggled with reconstructing intricate details. More recent research has turned to the high-dimensional feature space F, which successfully inverses the input image but loses much of the detail during editing. In this paper, we introduce StyleFeatureEditor -- a novel method that enables editing in both w-latents and F-latents. This technique not only allows for the reconstruction of finer image details but also ensures their preservation during editing. We also present a new training pipeline specifically designed to train our model to accurately edit F-latents. Our method is compared with state-of-the-art encoding approaches, demonstrating that our model excels in terms of reconstruction quality and is capable of editing even challenging out-of-domain examples. Code is available at \url{https://github.com/AIRI-Institute/StyleFeatureEditor}.
\end{abstract}

\section{Introduction}
\label{sec:intro}

In recent years, GANs \cite{goodfellow2014generative} have achieved impressive results in image generation, which has led to their use in a wide variety of computer vision tasks. One of the most successful models is StyleGAN \cite{karras2019style, karras2020analyzing, karras2021alias, karras2020training}, which not only has a high quality of generation, but also a rich semantic latent space. In this space, we can control different semantic attributes of the generated images by changing their latent code \cite{abdal2020image2stylegan++}. However, to apply this editing technique to real images, we must be able to find their internal representation in the StyleGAN latent space. This problem is called GAN inversion \cite{zhu2016generative}, and although it is well studied and many approaches have been proposed \cite{abdal2020image2stylegan++, zhu2020domain, richardson2021encoding, tov2021designing, alaluf2022hyperstyle, dinh2022hyperinverter,  yao2022feature, pehlivan2023styleres}, it is still an open problem to develop a method that simultaneously satisfies three requirements: high-quality reconstruction, good editability, and fast inference. Our work is dedicated to the development of such a method. 

Existing GAN inversion approaches can be divided into two groups: optimization-based and encoder-based. Optimization-based methods \cite{abdal2019image2stylegan, abdal2020image2stylegan++} learn a latent representation for each input image that best reconstructs that image. This results in good inversion quality, but such over-fitted latent codes may deviate from the original distribution of the latent space, resulting in poor editing. While there are approaches that improve the quality of editing by fine-tuning the generator itself for a given image \cite{roich2022pivotal}, this does not address the main drawback of such methods, which is that the inversion is too long, making them impractical to use in real-time applications. In contrast, 
more practical 
encoder-based methods \cite{richardson2021encoding, tov2021designing} allow us to obtain a latent representation of the input image in a single network pass. However, 
with these approaches, it is 
more difficult to achieve both high 
quality and good editability at the same time.   This is the so-called distortion-editability trade-off \cite{tov2021designing}. 
Inversion
quality and editability are directly related to the dimensionality of the latent space in which we encode the input image. 
In low-dimensional $W$ and $W^+$ spaces, we will get low reconstruction quality but high editability, because the low dimensionality of the latent code is a good regularizer that keeps it in the StyleGAN manifold. If we train the encoder to predict in the high-dimensional StyleGAN feature space $\mathcal{F}_k$, this will significantly increase the quality of the reconstructions at the expense of degraded editability, since in such a space it is easier to overfit to a particular image and escape the region in the latent space where semantic transformations work. Methods working in 
$\mathcal{F}_k$ \cite{wang2022high, yao2022feature, pehlivan2023styleres} try to challenge this problem by using additional transformations over the tensor $F_k$, but it is not completely solved. In particular, the editability problem is amplified when one increases the dimensionality of the $\mathcal{F}_k$ feature space by taking them from 
earlier
layers to improve the quality of reconstructions.

In this paper, we propose a framework that allows us to train an encoder in a high-dimensional $\mathcal{F}_k$ space that simultaneously achieves both excellent reconstruction quality and good editability. The main idea of our approach is to divide the training of our encoder into two phases. In the first, we train an encoder that predicts a latent code in $\mathcal{F}_k$ space with high resolution, which allows us to reconstruct images with high quality, but at the same time significantly reduces editability. 
To recover the editability 
, we introduce a second phase of training :
we propose to train a new Feature Editor module that task is to modify the feature tensor $F_k$ to obtain the target editing in image generation. 
The main difficulty in training this module is that we do not have a training data, where for each image there would be its edited versions. Therefore, we proposed to automatically generate such data using an encoder operating in $W^+$ space. That is, as training samples for Feature Editor, we take reconstructions of real images using a standard encoder with low inversion quality, but with good editability. And on this data we train the Feature Editor, which predicts $F'_k$ for the feature tensor $F_k$ of the input image, from which its edited version should be generated.

Thus, thanks to the proposed two-phase encoder learning framework, we are able to train an inversion model that has both high reconstruction quality, significantly better than the current state-of-the-art, and good editability. We conducted extensive experiments, demonstrating a significant improvement over state-of-the-art methods in the inversion task, and comparable results in the image editing. In particular, we have significantly improved the reconstruction metrics in terms of LPIPS and $L_2$ by more than a factor of 4 compared to StyleRes\cite{pehlivan2023styleres}, while the running time is equivalent to conventional encoder-based methods.

\section{Related Work}
\label{sec:rel_works}

\begin{figure*}[t]
\centering
\includegraphics[width=\linewidth]{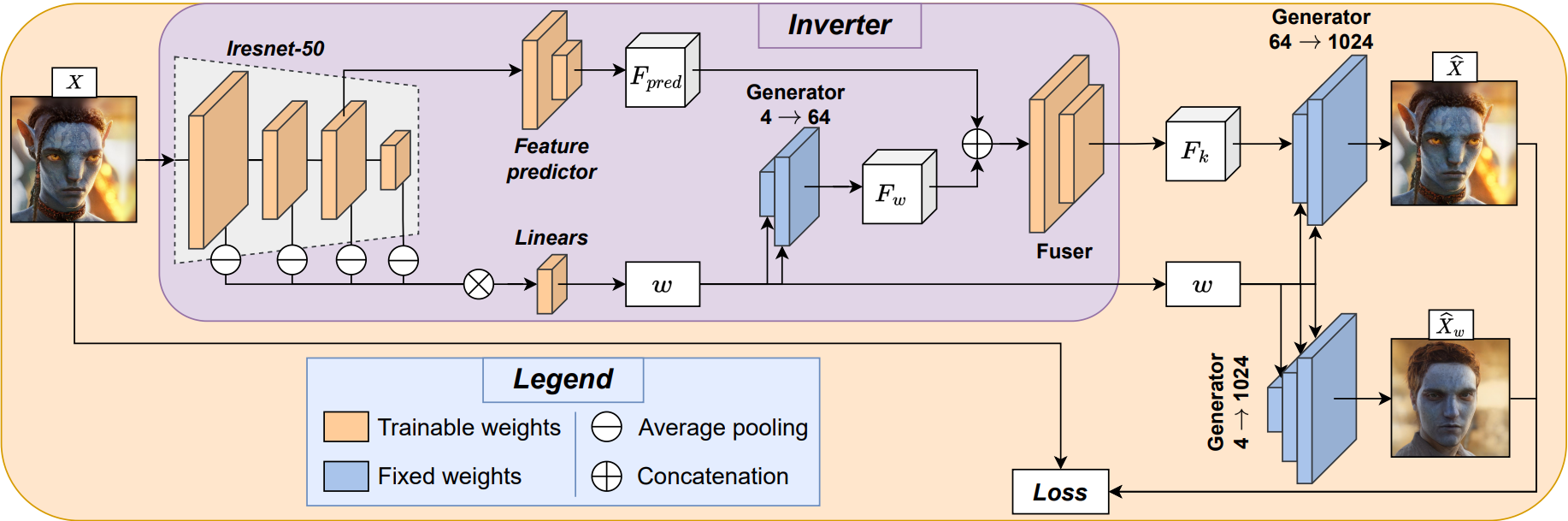}
\caption{\textbf{The Inverter training pipeline.} Input image $X$ is passed to Feature-Style-like backbone that predicts $w \in W^+$ and $F_{pred} \in \mathcal{F}_k$. Then $F_w = G(w_{0:k})$ is synthesized and passed with $F_{pred}$ to the Fuser that predicts $F_k$. Inversion $\widehat{X} = G(F_k, w_{k+1:N})$ is generated. Additional reconstruction $\widehat{X}_w = G(w_{0:N})$ is synthesized from w-latents only. Loss is calculated for pairs $(X, \widehat{X})$ and $(X, \widehat{X}_w)$.  \vspace{-0.4cm}} 
\label{fig:phase1}
\end{figure*}

\textbf{Latent Space Manipulation.} 
With the development of StyleGAN models \cite{karras2019style, karras2020analyzing, karras2021alias, karras2020training}, they started to be actively used for the task of image editing. Many methods have shown that by changing the latent code of an image in the latent space of StyleGAN, it is possible to change the semantic attributes of the image \cite{abdal2020image2stylegan++}. There are methods that find such directions using supervised approaches utilized attribute labelled samples or pre-trained classifiers \cite{abdal2021styleflow, goetschalckx2019ganalyze, shen2020interpreting, tewari2020stylerig}. Unsupervised methods do not use any kind of labelling \cite{harkonen2020ganspace, shen2021closed, voynov2020unsupervised, cherepkov2021navigating}, instead they, for example, perform PCA either in StyleGAN's feature space \cite{harkonen2020ganspace} or find directions in the weight space \cite{cherepkov2021navigating}. Other methods use a self-supervised learning approach \cite{jahanian2019steerability, plumerault2020controlling, spingarn2020gan}. And there are methods that utilize language-image models \cite{radford2021learning} to find desired edits guided by text \cite{patashnik2021styleclip, xia2021tedigan, gal2022stylegan}. To apply all these methods to real images, it is necessary first to encode images in StyleGAN's latent space.

\textbf{GAN Inversion.} 
The task of GAN inversion \cite{zhu2016generative} is to find the latent code for a real image, from which it can be generated by StyleGAN and the result has to be perceptually equal to the input image and can be edited by changing this latent code. Existing GAN inversion methods can be divided into two types: optimization-based methods \cite{abdal2019image2stylegan, abdal2020image2stylegan++, creswell2018inverting, zhu2020domain, zhu2021barbershop, roich2022pivotal, cao2023decreases, bhattad2023make} and encoder-based methods \cite{pidhorskyi2020adversarial, zhu2020domain, richardson2021encoding, tov2021designing, alaluf2022hyperstyle, dinh2022hyperinverter, alaluf2021restyle, wang2022high, yao2022feature, pehlivan2023styleres, hu2022style, liu2023delving, bai2022high}.

\textbf{Optimization-based methods.} Optimization methods find latent code by optimizing directly over the reconstruction losses. The first approaches \cite{abdal2019image2stylegan, abdal2020image2stylegan++, creswell2018inverting, zhu2020domain} performed optimization in $Z$/$W$/$W^+$ spaces. To improve the quality of the reconstruction, later methods proposed to optimize additionally in the StyleGAN feature space \cite{zhu2021barbershop}. Since the latent code can escape from the StyleGAN manifold during the optimization process and thus negatively affect the editability, it has been proposed to additionally fine-tune the generator weights for each image \cite{roich2022pivotal}. Although high reconstruction quality and good editability can be achieved with these approaches, the optimization process is too long, requiring up to several minutes for each image, which is not applicable for real-time interactive editing. 

\textbf{Encoder-based methods.} 
Encoder-based methods allow learning the mapping from the space of real images to the StyleGAN latent space in one or more passes through the neural network. Basically, these methods differ in the latent spaces they encode to. The first methods trained the mapping for the simplest $Z$, $W$, $W^+$ spaces \cite{pidhorskyi2020adversarial, zhu2020domain, richardson2021encoding, tov2021designing}, which gave good editability but low reconstruction quality. Next methods were proposed that additionally predicted changes in the generator weights using a hypernetwork to better reconstruct the input image \cite{alaluf2022hyperstyle, dinh2022hyperinverter}. This increased the quality of the reconstruction without sacrificing editability. There are also methods that propose to use multiple passes over the encoder to refine the details of the image during reconstruction \cite{alaluf2021restyle, alaluf2022hyperstyle}. The most successful methods train encoders for StyleGAN's feature space $\mathcal{F}_k$ \cite{wang2022high, yao2022feature, pehlivan2023styleres}. Such methods achieve the highest reconstruction quality among encoder-based methods, and are comparable to optimization-based methods. The main remaining problem is poor editability, since in such a high-dimensional latent space it is very easy to overfit the image and go out of the natural StyleGAN manifold. 

In our paper, we propose a framework that preserves the editability of an encoder trained in the StyleGAN's feature space $\mathcal{F}_k$, and achieves phenomenal reconstruction quality.

\section{Method}

\begin{figure*}[t]
\centering
\includegraphics[width=\linewidth]{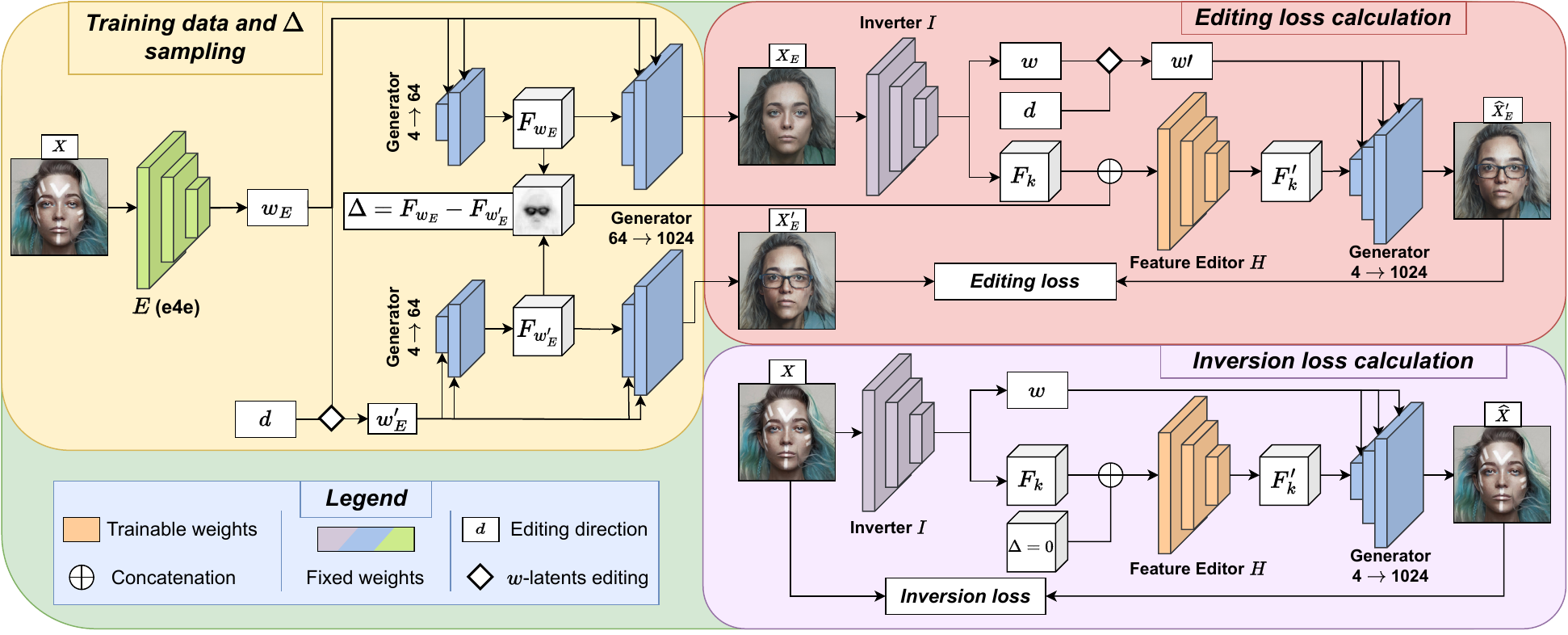}
    \caption{\textbf{The Feature Editor training pipeline and inference.} To obtain \textbf{editing loss}, one need to synthesize training samples: $X_{E}$ -- training input, and $X_{E}'$  -- training target. The pre-trained encoder $E$ takes the real image $X$ and predicts $w_{E} \in W^+$. Edited direction $d \in \mathcal{D}$ is randomly sampled, after which $w_E$ is edited to $w_E' = w_E + d$. Image $X_{E}$ and intermediate features $F_{w_{E}}$ are synthesized from $w_{E}$, while $X_{E}'$ and $F_{w_{E}'}$ are synthesized from $w_{E}'$ via generator $G$. $X_{E}$ is used as input and passed to frozen Inverter $I$ that predicts $F_k$ and $w$ that is edited to $w'$ according sampled $d$. Then $\Delta$ is calculated, and Feature Editor $H$ edits $F_k$ according $\Delta$. The edited reconstruction $\widehat{X}_{E}'$ is synthesized from $F_k'$ and $w_{k+1:N}'$. \textbf{Editing loss} is calculated between $X_{E}'$ and $\widehat{X}_{E}'$. To obtain the \textbf{inversion loss}, the real image $X$ is passed to $I$ that predicts $w$ and $F_k$, $F_k$ is edited to $F_k'$ by $H$ with $\Delta = 0$. The inversion $\widehat{X}$ is synthesized from $F_k'$ and $w_{k+1:N}$. The Inversion loss is calculated between $X$ and $\widehat{X}$. \textbf{Inference pipeline} is the same as synthesizing $\widehat{X}_{E}'$ but with the assumption that $I$ takes real image $X$ instead of $X_E$.  \vspace{-0.5cm}}
\label{fig:phase2}
\end{figure*}

\subsection{Overview}
\label{sec:back}
The goal of StyleGAN inversion methods is to find an internal representation of the input image in the StyleGAN latent space that contains as much information and detail as possible about the image itself, and at the same time allows editing it. This internal representation can be searched in different StyleGAN latent spaces, which have different properties. We can distinguish two main latent spaces that are considered in the StyleGAN inversion task, namely $W^+$ and $\mathcal{F}_k$. $W^+$ is the concatenation of $N$ vectors $w_1$, …, $w_N$, which are fed into each of the $N$ convolutional layers of StyleGAN. $\mathcal{F}_k$ is feature space, which is the combination of the $W^+$ space and the space of the feature outputs of the $k$-th convolutional layer of the StyleGAN.  

It is known that the representation of an image in $W^+$ space preserves few details, but allows good editing. In $\mathcal{F}_k$ space the situation is the opposite -- we can almost perfectly reconstruct the original image, but this representation is difficult to edit. The latest most advanced encoders Feature-Style \cite{yao2022feature} and StyleRes \cite{pehlivan2023styleres} work in $\mathcal{F}_k$ space, and to solve the editing problem, they offer their own techniques to transform the $F_k\in \mathcal{F}_k$ feature tensor during editing. But these techniques do not solve the problem completely. And it is exacerbated if the resolution of the $F_k$ feature tensor is increased. In this case, the quality of reconstructions improves significantly, but the editability completely vanishes.

In our work, we propose a way to edit  $F_k$ feature tensor that preserves high quality of the reconstruction with good editability. The basic idea is to train an additional module called Feature Editor,  which will transform the feature tensor $F_k$ in the right way for each edit. But to train Feature Editor, we will need a special training dataset, where for each image we need to have its edited versions. It is clear that it is very difficult  and expensive  to manually build such a dataset. Therefore, we generated this dataset using an encoder that operates in $W^+$ space. That is, for each real image from the dataset, we find its reconstruction in $W^+$ space, get its edited version, and use these two images to train our Feature Editor module. This approach allowed us to significantly improve the quality of edits, even for high resolutions of $F_k$. Further, we give more details about the architecture of StyleFeatureEditor and the training process.

\subsection{Architecture}

In this section, we describe StyleFeatureEditor, which consists of two parts: Inverter $I$ and Feature Editor $H$. The task of Inverter is to extract reconstruction features from the input image, while Feature Editor should transform these  features according to the information about the desired edit.
 
\textbf{Inverter.} $I$ consists of Feature-Style-like Encoder $I_{fse}$ and an additional module  $I_{fus}$ called Fuser. $I_{fse}$ consists of Iresnet50 backbone, Feature predictor and Linear layers (see Fig. \ref{fig:phase1}). First, the input image $X$ is passed to the backbone, which predicts 4 intermediate features, pools them to the same dimensionality, concatenates them, and maps to $w \in W^+$ by Linear layers. The third intermediate feature is also passed to Feature predictor that predicts $F_{pred} \in \mathcal{F}_k$:
 \begin{equation}
    \label{eq:inv_fe}
    (w, F_{pred}) = I_{fse}(X).
 \end{equation}

 Despite good inversion quality, Feature-Style Encoder fails to reconstruct fine details of the image, thus we increased the predicted feature tensor from $F_{pred} \in \mathcal{F}_5$ to $F_{pred} \in \mathcal{F}_9$ that increases its dimensionality from $\mathbb{R}^{512 \times 16 \times 16}$ to $\mathbb{R}^{512 \times 64 \times 64}$ respectively.

 To take into account the impact of $w_{0:k}$ we additionally synthesize output of the $k$-th generator layer $F_w = G(w_{0:k})$ via the StyleGAN2 generator $G$. $F_w$ then fused with predicted $F_{pred}$ by an additional module $I_{fus}$, which predicts $F_k \in \mathcal{F}_k$:
 \begin{equation}
 \label{eq:inv_fuse}
 F_k = I_{fus}(F_{pred}, F_w)
  \end{equation}
Thus, $I$ takes input image $X$ and predicts $w$ and $F_k$:
 \begin{equation}
    \label{eq:inv}
    (w, F_{k}) = I(X).
 \end{equation}
after then, the reconstructed image $\widehat{X}$ is synthesized from $F_k$ and $w_{k+1}, \dots, w_N$:
\begin{equation}
    \label{eq:synt_inv}
    \widehat{X} = G(F_k, w_{k+1:N}).
\end{equation}
It is also possible to synthesize image $\widehat{X}_w = G(w)$ from w-latents only, which we use during training. 

\textbf{Feature Editor.} The predicted feature tensor $F_k$ contains much of the input image information, which allows even finer image details to be reconstructed. However, if we do not transform $F_k$ during editing, artefacts may appear or editing may not work at all. Therefore, we propose an additional Feature Editor module $H$ that transforms predicted $F_k$ according to the desired edit. In order for $H$ to understand what to change, it is necessary to have information $\Delta$ about which regions $F_k$ need to be edited. To obtain such information, we propose to use a pre-trained encoder $E$ in $W^+$ space that is capable of good editing (we use pre-trained e4e encoder \cite{tov2021designing}).

$E$ takes input image $X$ and predicts $w_{E} = E(x) \in W^+$, which is edited to $w_{E}' = w_{E} + d$ by editing direction $d$. The outputs of the k-th generator layer $F_{w_{E}}$ and $F_{w_{E}'}$ are synthesized from $w_{E}$ and $w_{E}'$ respectively. Difference between $ F_{w_{E}}$ and $ F_{w_{E}'}$ contains information about edited regions:
\begin{equation}
    \label{eq:delta}
    \Delta = F_{w_{E}} - F_{w_{E}'}.
\end{equation} 
After gaining $\Delta$, $H$ transforms $F_k$ to $F_k'$ according $\Delta$:
\begin{equation}
    \label{eq:fe}
    F_k' = H(F_k, \Delta).
\end{equation} 
The edited image $\widehat{X}'$ is synthesized from $F_k'$ and $w_{k+1}', \dots, w_N'$, where $w'$ is edited $w$ (see Fig. \ref{fig:phase2}):
 \begin{equation}
    \label{eq:synt_ed}
    \widehat{X}' = G(F_k', w_{k+1:N}').
\end{equation}

To sum up, the inference  pipeline of StyleFeatureEditor during editing consists of predicting $w$ and $F_k$ (Eq. \ref{eq:inv}), editing $w$ to $w' = w + d$, computing $\Delta$ according Eq. \ref{eq:delta}, editing $F_k$ (Eq. \ref{eq:fe}) and synthesizing edited image (Eq. \ref{eq:synt_ed}). Inversion assumes the same pipeline, but with $\Delta = 0$.

\begin{figure*}[!htbp]
    \centering 
    \large
    \begin{tabular}{cc}
    
    \rotatebox{90}{\hspace{2.5mm} Inversion}  & \includegraphics[width= 0.9\linewidth]{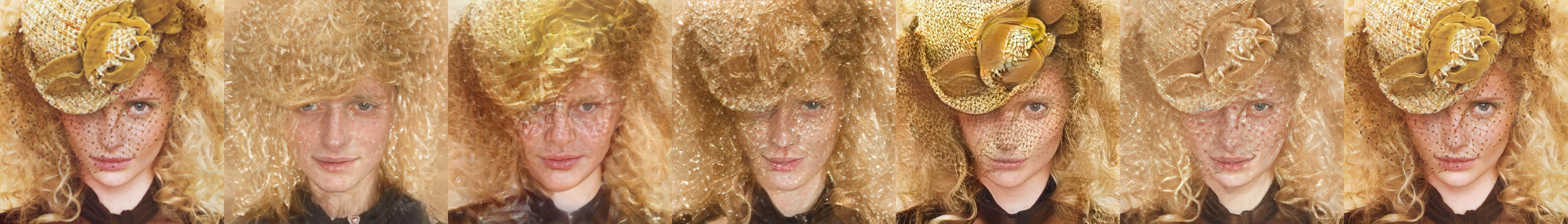}\\

   \rotatebox{90}{\hspace{3.5mm} Glasses}  & \includegraphics[width= 0.9\linewidth]{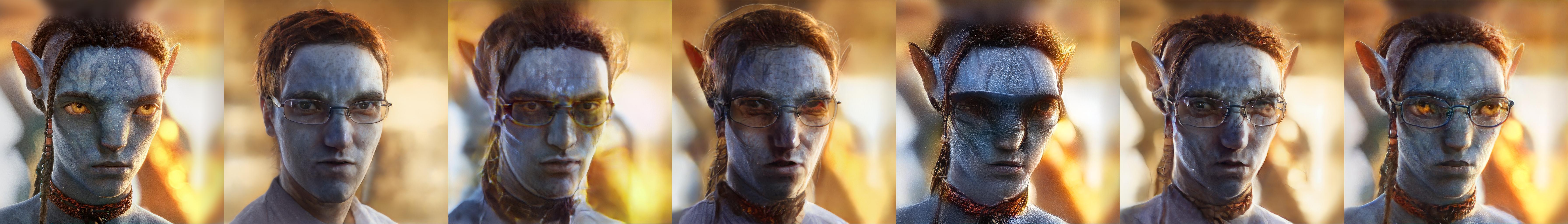}\\
   
   \rotatebox{90}{\hspace{2mm} Black hair}  & 
   \includegraphics[width= 0.9\linewidth]{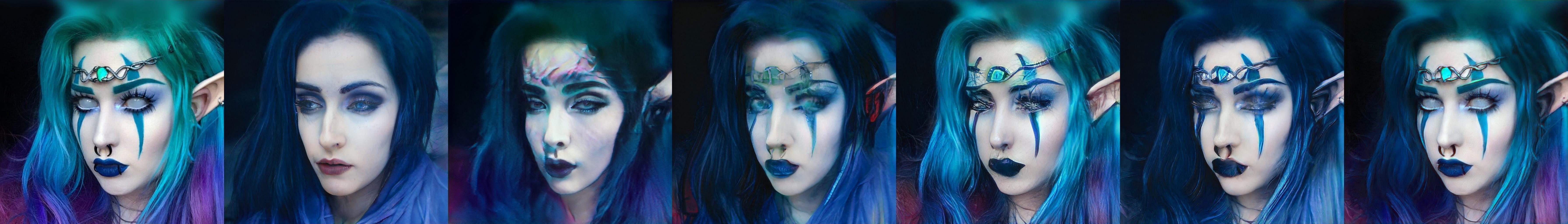}\\ 

   \rotatebox{90}{\hspace{5mm} Bobcut}  & 
   \includegraphics[width= 0.9\linewidth]{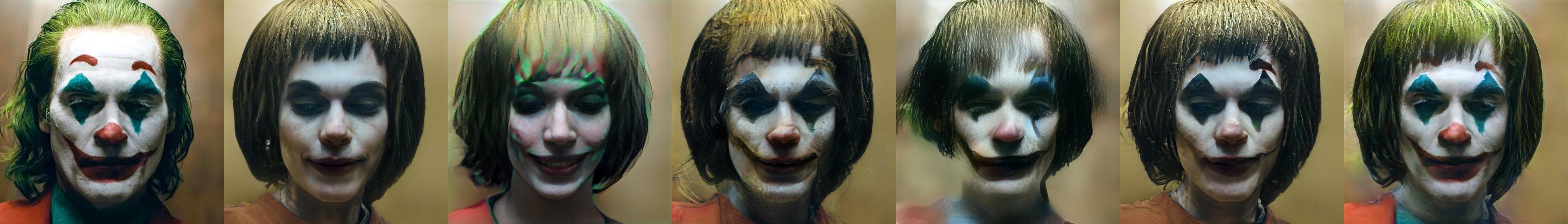}\\

    \end{tabular}
    \begin{tabular*}{0.85\linewidth}{@{\extracolsep{\fill}}ccccrrr}
     \hspace{6mm} Input &  \hspace{7mm} e4e &  \hspace{4mm}Hyperinverter \hspace{2mm} & HFGI \hspace{3mm} & \hspace{8mm}FS & \hspace{6mm}StyleRes & \hspace{4mm}SFE (ours)
    \end{tabular*}
    \caption{Visual comparison of our method with previous encoder-based approaches on face domain. Row 1 represents the inversion, row 2 -- the addition of glasses, row 3 -- the darkening of the hair colour, row 4 -- the changing of the hairstyle.}
    \label{fig:edit_face}
\end{figure*}

\subsection{Training Inverter (Phase 1)}
This section is related to training Inverter $I$ to reconstruct source images. The pipeline of phase 1 is presented in Fig. \ref{fig:phase1}. 

The source image $X$ is passed to $I$ which predicts $w, F_k = I(X)$, where $w \in \mathbb{R}^{N \times 512}$ and $F_k \in \mathbb{R}^{512 \times 64 \times 64}$. Then the generator $G$ synthesizes 
 $\widehat{X} = G(F_k, w_{k+1:N})$ -- reconstruction of the input image $X$. The loss function $\mathcal{L}_{phase1}$ is calculated between $X$ and $\widehat{X}$. In addition, to force information flow not only through feature space $\mathcal{F}_k$, we also calculate $\mathcal{L}_{phase1}$ for reconstruction $\widehat{X}_w  = G(w)$ obtained from $w$-latents only.

The loss function $\mathcal{L}_{phase1}$ consists of two equal parts: the image loss $\mathcal{L}_{im}$ applied to both $(X, \widehat{X})$ and $(X, \widehat{X}_w)$, and the regularization $\mathcal{L}_{reg}$ for constraining the norm of $F_k$ tensor. $\mathcal{L}_{im}$ is calculated as a weighted sum of per-pixel loss $\mathcal{L}_2$, perceptual LPIPS loss $\mathcal{L}_{lpips}$ \cite{zhang2018unreasonable}, identity-based similarity loss (ID) $\mathcal{L}_{id}$ \cite{richardson2021encoding} by utilizing a pre-trained network (ArcFace \cite{deng2019arcface} for the face domain and ResNet-based \cite{tov2021designing} for non-facial domains), adversarial loss $\mathcal{L}_{adv}$ by employing a pre-trained StyleGAN discriminator $D$ which we fine-tune during training. As the regularization loss, we use $\mathcal{L}_{reg} = \|F_k\|_2$. So, the total loss $\mathcal{L}_{phase1}$ is calculated as:

\begin{gather}
\mathcal{L}_{im} = \mathcal{L}_2 + \lambda_{lpips}\mathcal{L}_{lpips} + \lambda_{id}\mathcal{L}_{id} + \lambda_{adv}\mathcal{L}_{adv}, \\
\label{eq:loss_ph1}
\mathcal{L}_{phase 1} = \mathcal{L}_{im} + \lambda_{reg}\mathcal{L}_{reg},
\end{gather}
where $\lambda_{lpips} = 0.8, \lambda_{id} = 0.1, \lambda_{adv} = 0.01, \lambda_{reg} = 0.01.$

\subsection{Training Feature Editor (Phase 2)}

The goal of this phase is to train the Feature Editor $H$ to edit  $F_k$. The training pipeline of this phase is available in Fig. \ref{fig:phase2}. In this phase, we assume that $I$ is already trained, so we froze its weights and train only $H$ weights. 

For this purpose it is necessary to have a dataset consisting of pairs ($X$, $X'$), where $X'$ is the edited version of the image $X$, but it is difficult to collect such data manually. Therefore, we propose to use a pre-trained encoder $E$ in $W^+$ space suitable for editing to generate such data. $E$ takes input image $X$ and predicts $w_{E}$, it is edited with specified editing direction $d$ to $w_{E}' = w_{E} + d$, after which images $X_{E}$ and $X_{E}'$ are synthesized from $w_{E}$ and $w_{E}'$ respectively. 

During this phase, we fix a set of 13 editing directions $\mathcal{D}$ used in training (more details in Appendix \ref{apend:train}). The pipeline of training $H$ using synthetic data is:
\begin{enumerate}
    \item Pass $X$ to $E$ to obtain $w_{E}$ and $w_{E}' = w_{E} + d$ for the editing direction $d$  randomly sampled from $\mathcal{D}$.
    \item Synthesize images $X_{E}$, $X_{E}'$ and feature tensors $F_{w_{E}}, F_{w_{E}'}$ from $w_{E}$ and  $w_{E}'$ respectively. 
    \item Calculate $\Delta = F_{w_{E}} - F_{w_{E}'}$.
    \item Compute $(w, F_k) = I(X_{E})$.
    \item Obtain the edited tensor $F_k' = H(F_k, \Delta)$.
    \item Synthesize $\widehat{X}_{E}' = G(F_k', w_{k+1:N}')$ -- the edited reconstruction.
    \item Calculate the loss  between $\widehat{X}_{E}'$ and $X_{E}'$.
\end{enumerate}

However, if $H$ is trained only on synthetic images, the reconstruction quality for real images may degrade. To solve this problem, we propose to train $H$ not only on editing, but also on the classical inversion task. The training pipeline is the same, but for inversion we use a real image $X$ as input and assume $\Delta$ = 0. $X$ is passed to $I$, which predicts $F_k$ and $w$ (Eq. \ref{eq:inv}), $\Delta = 0$ and $F_k$ goes to the Feature editor which predicts $F_k'$ and reconstruction $\widehat{X}$ is synthesised assuming $w' = w$ (Eq. \ref{eq:synt_ed}). The loss is calculated between $X$ and its reconstruction $\widehat{X}$.

For this phase we used $\mathcal{L}_2$, $\mathcal{L}_{lpips}$ and $\mathcal{L}_{id}$ for both inversion and editing tasks with coefficients from phase 1. For inversion task we additionally use adversarial loss $\mathcal{L}_{adv}$:
\begin{gather}
    \label{eq:loss_ed}
    \mathcal{L}_{edit} = \mathcal{L}_2 + \lambda_{lpips}\mathcal{L}_{lpips} + \lambda_{id}\mathcal{L}_{id}, \\
\label{eq:loss_inv}
    \mathcal{L}_{inv} = \mathcal{L}_2 + \lambda_{lpips}\mathcal{L}_{lpips} + \lambda_{id}\mathcal{L}_{id} + \lambda_{adv}\mathcal{L}_{adv}.
\end{gather}
The general loss $\mathcal{L}_{phase2}$ for phase 2 is calculated as:
\begin{equation}
    \label{eq:loss_ph2}
    \mathcal{L}_{phase2} = \mathcal{L}_{edit}(X_0', \widehat{X}_0') + \mathcal{L}_{inv}(X, \widehat{X})
\end{equation}
During training we fixed a set of 13 editing directions $\mathcal{D}$, however SFE is capable of generalising to new directions without any retraining. Furthermore, $\mathcal{D}$ can be restricted while SFE's editing abilities remain good on both: seen and unseen directions (see Ablation Study \ref{ablat}, Appendix \ref{app:edit}). This can be explained by the fact that $\Delta$ (which contains almost all editing information) of even one direction will be very different for different images. Therefore, during training, $H$ does not learn specific directions, but generalizes to gather information from $\Delta$. Since $\Delta$ depends only on the edited $w$-latents obtained from E (e4e), our method is able to apply any editing applicable to E (e4e).

More training and architecture detail available in the Appendix \ref{apend:train}, \ref{apend:arch}.

\section{Experiments}
\label{sec:exps}
\subsection{Experiment set-up}

 In our experiments for face domain, we used FFHQ \cite{karras2019style} image dataset for training and official test part of Celeba HQ dataset \cite{karras2018progressive} for inference. For the car domain, we used train part of Stanford Cars \cite{6755945} for training and test part for evaluation. For test editings we used InterfaceGAN\cite{shen2020interpreting} and Stylespace\cite{Wu2020StyleSpaceAD} for both face and car domains,  StyleClip\cite{patashnik2021styleclip} and GANSpace\cite{harkonen2020ganspace} for face domain. To extract $\Delta$ and sample images for editing loss calculation during training phase 2, we used pre-trained e4e \cite{tov2021designing} as $E$. For the inversion calculation, we used our full pipeline including both $I$ and $H$, assuming $\Delta = 0$ as in Fig. \ref{fig:phase2}.

We compare our method with state-of-the-art encoder approaches such as e4e\cite{tov2021designing}, psp\cite{richardson2021encoding}, StyleTransformer\cite{hu2022style}, ReStyle\cite{alaluf2021restyle}, PaddingInverter\cite{bai2022high}, HyperInverter\cite{dinh2022hyperinverter}, Hyperstyle\cite{alaluf2022hyperstyle}, HFGI\cite{wang2022high}, Feature-Style\cite{yao2022feature}, StyleRes\cite{pehlivan2023styleres}  and optimisation-based PTI\cite{roich2022pivotal}. We used author's original checkpoints, but in car domain, some of them are not public. We train Feature-Style on Stanford Cars by using authors code and omitting models without official checkpoints.

\begin{figure*}[htbp]
    \centering 
    \large
    \begin{tabular}{cc}

   \rotatebox{90}{ Inversion}  &
   \includegraphics[width= 0.9\linewidth]{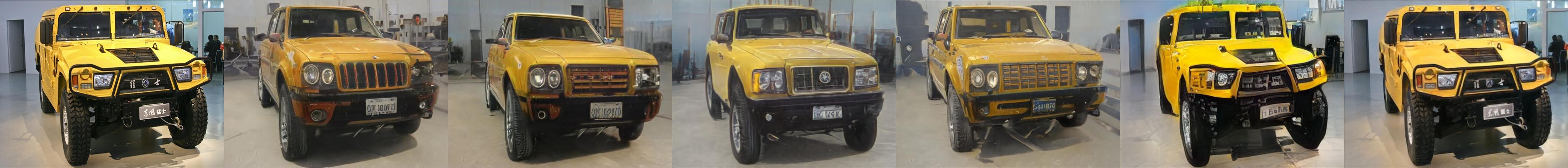}\\ 

   \rotatebox{90}{\hspace{3.5mm} Grass}  &
   \includegraphics[width= 0.9\linewidth]{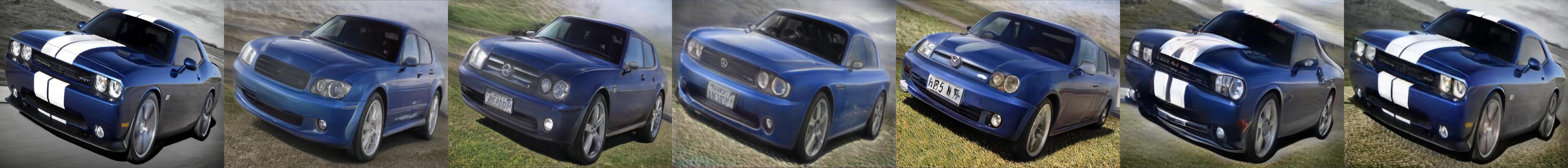}\\

    \rotatebox{90}{\hspace{3.5mm} Color}  &
   \includegraphics[width= 0.9\linewidth]{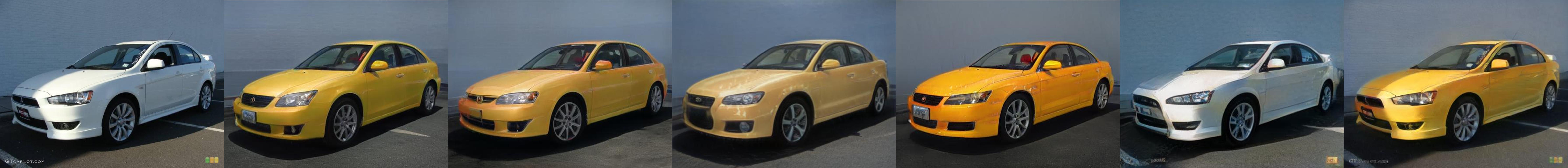}\\

    \end{tabular}
    \begin{tabular*}{0.85\linewidth}{@{\extracolsep{\fill}}ccccccc}

     \hspace{4mm} Input &  \hspace{12mm} e4e &  \hspace{7mm} ReStyle & \hspace{0mm} StyleTrans \hspace{1mm} &  HyperStyle & \hspace{5mm} FS & \hspace{6mm} SFE (ours) \hspace{12mm} \\
    \end{tabular*}
    \caption{Additional visual comparison of our method with previous encoder-based approaches in the car domain. Row 1 represents the inversion, Row 2 -- the addition of grass, Row 3 -- the change in car colour.}
    \label{fig:edit_car}
\end{figure*}

\renewcommand\arraystretch{1}
\renewcommand{\tabcolsep}{3pt}
\begin{table*}[t]
    \centering
    \caption{Quantitative comparison results for inversion quality and editing abilities on face domain. To measure inversion we report LPIPS, $L_2$, MS-SSIM and FID calculated on Celeba HQ test set. To measure editing abilities, we used FID as described in \ref{quant}.We also measured the time required to edit a single image on a single TeslaV100.} 
    \label{tab:metr_face}
    \tiny
    \resizebox{1.0\linewidth}{!}{
    \begin{tabular}{lcccccccc}
        \toprule \vspace{-0.05cm} & \multicolumn{4}{c}{\textbf{Inversion quality}} & \multicolumn{3}{c}{\textbf{Editing quality}} & \vspace{-0.05cm}\\
        \cmidrule(lr){2-5}                  
        \cmidrule(lr){6-8}
        
        Model &  LPIPS $\downarrow$ &     L2 $\downarrow$ &     FID $\downarrow$ &     MS-SSIM $\uparrow$  & Smile (-) & Glasses (+) & Old (+) & Time (s) \vspace{-0.05cm}
        \\
            \midrule[0.01em] e4e\cite{tov2021designing} &  0.199 &  0.047 &   28.971 & 0.625 &  51.245 & 119.437 & 68.463  &  0.034\\
               pSp\cite{richardson2021encoding} &  0.161 &  0.034 &    25.163 &  0.651  & 46.220 & 105.740 & 67.505 & 0.034\\
               StyleTransformer\cite{hu2022style} &  0.158 &  0.034 &  22.811 & 0.656   & 32.936 & 81.031 & 67.250 & 0.032\\
                   ReStyle\cite{alaluf2021restyle} &  0.130 &  0.028 & 20.664 & 0.669 & 36.365 & 87.410 & 56.025 & 0.138\\
                   Padding Inverter\cite{bai2022high} &  0.124 &  0.023 & 25.753 &  0.672  & 42.305 & 98.719 & 62.283 & 0.034 \vspace{-0.05cm}\\
                \midrule[0.01em]  HyperInverter\cite{dinh2022hyperinverter} &  0.105 &  0.024 &  16.822 & 0.673 & 41.201 & 93.723 & 65.282 & 0.105 \\
                HyperStyle\cite{alaluf2022hyperstyle} &  0.098 &  0.022 &  20.725 & 0.700 & 34.578 & 86.764 & 49.267 & 0.275 \vspace{-0.05cm}\\
                \midrule[0.01em] HFGI\cite{wang2022high} &  0.117 &  0.021 &  15.692 & 0.721 & 27.151 & 77.213 & 51.489 & 0.072\\
          Feature-Style\cite{yao2022feature} &  \underline{0.067} &  0.012 & 10.861 &  0.758 & 26.034 & 85.686 & 56.050 & 0.038  \\
          
          StyleRes\cite{pehlivan2023styleres} &  0.076 &  0.013 &  \underline{8.505} &  \underline{0.797} & \underline{24.465} & \textbf{73.089} & \underline{43.698} &  0.063 \vspace{-0.05cm}\\
          \midrule[0.01em]
          PTI\cite{roich2022pivotal}  &  0.085 &  \underline{0.008} & 14.466 &  0.781 & 28.302 & 78.058 & 44.856 & 124 \vspace{-0.05cm}\\
          \midrule[0.01em]
           SFE (ours) &  \textbf{0.019} &  \textbf{0.002} &  \textbf{3.535} &  \textbf{0.922}  & \textbf{24.388} & \underline{73.098} & \textbf{41.677} & 0.070 \vspace{-0.05cm}\\
        \bottomrule
    \end{tabular}
}
\end{table*}

\begin{table}[t]
    \centering
    \caption{Additional quantitative comparisons on the Stanford Cars dataset. We do not provide a calculation of editing ability, as the test set does not have the required markup.} 
    \label{tab:metr_cars}
    \tiny
    \resizebox{1.0\linewidth}{!}{
    \begin{tabular}{lccccc}
         \toprule
        Model &  LPIPS $\downarrow$ &     L2 $\downarrow$ &     FID $\downarrow$ & \vspace{-0.05cm}\\
        \midrule
                 e4e\cite{tov2021designing} &  0.325 &  0.122 &   13.397 \\
            
                   ReStyle\cite{alaluf2021restyle} &  0.306 &  0.102 & 13.008 \\
                   StyleTransformer\cite{hu2022style} & 0.276 &  0.092 &  10.644  \\

                   HyperStyle\cite{alaluf2022hyperstyle} & 0.287 &  0.080 &  8.044  \\

          Feature-Style\cite{yao2022feature} &  \underline{0.147} &  \underline{0.045} & \underline{7.180} \\
          
          SFE (ours) &  \textbf{0.039} &  \textbf{0.004} &  \textbf{4.035} \vspace{-0.05cm}\\
        \bottomrule
    \end{tabular}
}
\end{table}

\begin{figure*}[htbp]
    \centering 
   \large
    \begin{tabular}{c}    
   \includegraphics[width= 1\linewidth]{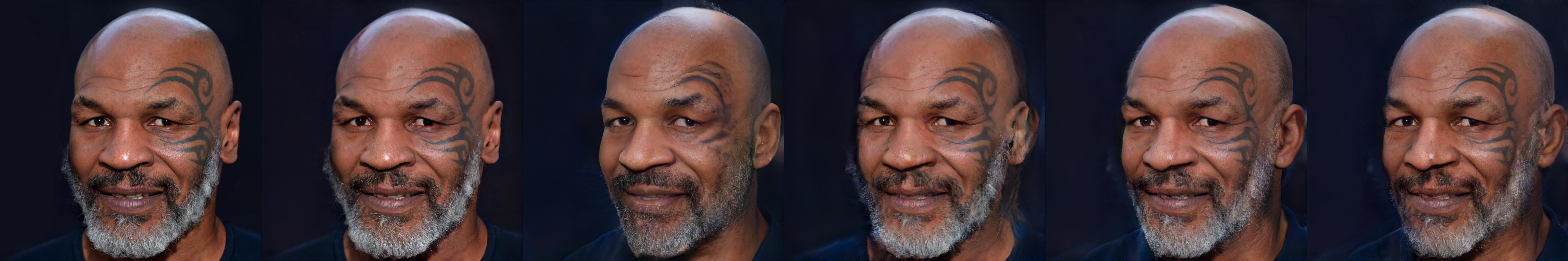}\\
   
    \end{tabular}
    \begin{tabular*}{1\linewidth}{@{\extracolsep{\fill}}cccccc}
      \hspace{10mm}Input & \hspace{4mm} W/o $H$ & \hspace{2mm} $\mathcal{F}_9 \rightarrow \mathcal{F}_5$   & W/o $E$ & $\mathcal{D}_{small}$ & Final \hspace{7mm} \\
    \end{tabular*}
    \caption{Ablation study. Visual representation of outputs of different ablations (described in \ref{ablat}) during pose rotation.}
    \label{fig:ablat}
    \vspace{-0.2cm}
\end{figure*} 

\begin{table}[t]
    \centering
    \caption{ Ablation study. Quantitative comparison of different ablations (described in \ref{ablat}). We calculated all metrics on the test part of Celeba HQ. To measure editing quality, we calculated FID as described in \ref{quant}.} 
    \vspace{-0.05cm}
    \label{tab:alat}
    \scriptsize
    \resizebox{1.0\linewidth}{!}{
    \begin{tabular}{lccccccc}
        
        \toprule \vspace{-0.05cm} & \multicolumn{3}{c}{\textbf{Inversion}} & \multicolumn{2}{c}{\textbf{Editing}} \\
        \cmidrule(lr){2-4}                  
        \cmidrule(lr){5-6}
        
        Model &  LPIPS $\downarrow$ &     L2 $\downarrow$ &     FID $\downarrow$    & Smile (-) & Old (+) \vspace{-0.05cm}\\
        \midrule
          Final model &  \underline{0.019} &  \underline{0.0017} &  \underline{3.535}  & \textbf{24.388} &  \textbf{41.677}\\
          \midrule[0.01em]
    
          W/o  $H$ &  \textbf{0.016} &  \textbf{0.0013} &  \textbf{2.975}  & 28.149 & 54.621 \\
          W/o Fuser  & 0.023 & 0.0019 & 4.239  & 26.410  & \underline{42.121}\\
          W/o inv loss & 0.037 & 0.0027 & 5.101  & 26.179 &  42.361\\
          W/o $E$ (e4e) & 0.021 & 0.0024 & 3.829 & 24.941 &  44.398\\
           $\tiny{F_9 \rightarrow F_5}$  & 0.064 & 0.0089 & 7.915 &  25.933  & 43.317 \\
          $\mathcal{D}_{small}$ & 0.021 & 0.0019 & 3.842  & \underline{24.548} &  42.317 \\
        \bottomrule
    \end{tabular}
}
\vspace{-0.5cm}
\end{table}

\subsection{Qualitative evaluation}
 To demonstrate the performance of our method, in Figure \ref{fig:edit_face} we compare it with previous approaches on several hard out-of-domain examples. Our approach not only reconstructs more detail than previous ones, but also preserves it during editing. For example, in the first row, our method accurately reconstructs woman's hat while others smooth it out.  In the second row, our method preserves the yellow eye colour while editing the eye zone.  In rows 3 and 4, it is evident that our approach is better at reconstructing difficult make-up and preserving the colours of the source image. \vspace{-0.09cm}

Additionally, in Figure \ref{fig:edit_car} we show comparison of our method on car domain. In the first row, our method even manages to reconstruct the original shape of a car when the others do not.  Moving on to the second row, our method most accurately reconstructs the outline and white lines of the original car, while FS Encoder distorts them. Apart from our approach in the third row, FS Encoder is the only one that can reconstruct the shadow on the car, but it fails in changing car colour.

\subsection{Quantitative evaluation} \label{quant}
 To evaluate the effectiveness of the inversion technique, two key aspects can be examined. Firstly, the accuracy of the inversion, which refers to the degree to which the method is able to reconstruct the details of the original image. Second, the editability -- how well the inverted image can be edited.The comparison in both aspects on CelebA-HQ dataset is presented in Table \ref{tab:metr_face}.

To measure quality of the inversion details, we used LPIPS \cite{zhang2018unreasonable}, $L_2$ and MS-SSIM \cite{1292216}. Additionally, we determined realism of the synthesized images by measuring distance between distributions of real and inverted images using FID \cite{Heusel2017GANsTB}. Our method outperformed all previous approaches. The most notable difference was seen in LPIPS and $L_2$, indicating that our method is capable of extremely fine detail reconstruction. We also tested our method in the domain of cars on the Stanford Cars dataset presented in Table \ref{tab:metr_cars}, which confirms the results described above.

It is challenging to accurately estimate the quality of the editing numerically in the absence of target images. To perform these calculations, we use the technique proposed in \cite{pehlivan2023styleres}. We determine the attribute to be edited, then, based on the Celeba HQ markup, we divide the test dataset into images $A$ with and $B$ without this attribute. Next, we apply the method to $B$ to add this attribute and synthesize $B'$. The FID between $B$ and $B'$ demonstrates the effectiveness of the technique for editing this attribute. We provide experiments with 3 attributes: removing smile, adding glasses and increasing age. 

The results show that our method not only inverts well, but is also comparable to the current state-of-the-art StyleRes in terms of editing capabilities. Furthermore, our method requires only 0.066 seconds to edit a single image on the TeslaV100, far outperforming optimisation-based PTI and matching previous encoder-based methods in terms of inference speed.

\subsection{Ablation Study} \label{ablat}
 \vspace{-0.1cm}
 To ensure the importance of each component in the proposed pipeline, we conducted several ablation experiments. We present the quantitative results of these experiments in Table \ref{tab:alat} and visual representations in Figure \ref{fig:ablat}. 
 
 First, we tried to discard $H$ and use only $I$ as in training phase 1. Despite a small increase in the inversion metrics, the edits stopped working, proving the significance of $H$. We also tried an architecture without Fuser $I_{fus}$ (which refers to the case where $F_k = F_{pred}$) and an experiment where the inversion loss is omitted during the second training phase. Both of these experiments resulted in a drop in reconstruction quality that is difficult to detect at low resolution and only visible at high resolution. The fourth experiment was related to omitting $E$ and predicting features for $\Delta$ from $w$ obtained from $I$. The predicted $w$ is much less editable than $w_{E}$ from e4e, leading to artefacts during editing (Figure \ref{fig:ablat}) and showing that $E$ should be well editable. 
 
 We also attempted to train our pipeline with a lower predicted feature dimensionality. We reduced the predicted $F_k$ from $k = 9$ to $k = 5$, which is the dimensionality of the Feature Space Encoder. Despite the significant decrease in inversion quality, this approach is still capable of good editing, unlike Feature-Style. During the last ablation, we reduced the number of editing directions in $\mathcal{D}$ from 13 to 6 in the second training phase. The reduced $\mathcal{D}_{small}$ consists of Age, Afro, Angry, Face Roundness, Bowlcut Hairstyle and Blonde Hair. Despite a slight decrease in metrics, our method is still able to edit directions that were not used during training, as shown in Figure \ref{fig:ablat}.
 \vspace{-0.2cm}

\section{Conclusion}
\vspace{-0.2cm}
In this paper, we have demonstrated StyleFeatureEditor -- a novel approach to image editing via StyleGAN inversion and introduced a new technique for training it. Even for challenging out-of-domain images, we have achieved a reconstruction quality that makes it almost impossible to tell the difference between the real and synthetic images with the naked eye. Thanks to Feature Editor, our method is not only able to reconstruct finer facial details, but also preserves most of them during editing.

\section{Acknowledgments}
The analysis of related work in sections 1 and 2 were obtained by Aibek Alanov with the support of the grant for research centres in the field of AI provided by the Analytical Center for the Government of the Russian Federation (ACRF) in accordance with the agreement on the provision of subsidies (identifier of the agreement 000000D730321P5Q0002) and the agreement with HSE University No. 70-2021-00139.
This research was supported in part through computational resources of HPC facilities at HSE University.

{
    \small
    \bibliographystyle{ieeenat_fullname}
    \bibliography{main}
}

\clearpage
\setcounter{page}{1}
\maketitlesupplementary

\section{Training details}\label{apend:train}
The training of the StyleFeatureEditor consists of two phases: Phase 1 -- training of the Inverter and Phase 2 -- training of the Feature Editor. A batch size of 8 is used for both phases. We used Ranger optimiser with a learning rate of 0.0002 to train each part of our model, and Adam optimiser with a learning rate of 0.0001 to train the Discriminator.

\textbf{Phase 1.} During this phase, we duplicate the batch of source images $X$ and synthesize the reconstruction $\widehat{X}$ and the reconstruction from w-latents only $\widehat{X}_w$ for the same images. 
The loss is computed for both pairs $(X, \widehat{X}), (X, \widehat{X}_w)$ and consists of $L_2$, LPIPS, ID, adversarial loss and regularization loss for predicted feature tensor $F_k$ with corresponding coefficients $\lambda_{loss}$. We used $\lambda_{L_2} = 1, \lambda_{lpips} = 0.8, \lambda_{id} = 0.1$ for face domain and $\lambda_{id} = 0.5$ for car domain, $\lambda_{adv} = 0.01, \lambda_{reg} = 0.01$. We start applying adversarial loss and training the discriminator only after 14'000 steps. The full training duration of the first phase is 37'500 steps.

\textbf{Phase 2.} As in phase 1, the batch of source images $X$ is duplicated and the same images are used for both inversion and editing loss. First, training samples $X_E$ and $X_E'$ are synthesized, then $X_E$ is passed through StyleFeatureEditor which tries to reconstruct and edit it to $\widehat{X}_E'$, the editing loss is calculated between $X_E'$ and $\widehat{X}_E'$. For inversion loss, the reconstruction $\widehat{X}$ is synthesized for the same images used for sampling $(X_E, X_E')$. The inversion loss is calculated between $X$ and $ \widehat{X}$. For both losses we use $L_2$, LPIPS and ID with the corresponding coefficients $\lambda_{L_2} = 1, \lambda_{lpips} = 0.8, \lambda_{id} = 0.1$ for face domain and $\lambda_{id} = 0.5$ for car domain. For inversion loss we additionally apply adversarial loss with $\lambda_{adv} = 0.01$. The duration of this phase is 20'000 steps.

During the second phase, we fix a set $\mathcal{D}$ of possible editing directions that we apply to compute the editing loss. $\mathcal{D}$ consists of InterfaceGAN\cite{shen2020interpreting} directions ("Age", "Smile", "Pose Rotation", "Glasses", "Make-up"), GANSpace\cite{harkonen2020ganspace} direction "Face Roundness", StyleClip\cite{patashnik2021styleclip} directions ("Afro", "Angry", "Bobcut Hairstyle", "Mohawk Hairstyle", "Purple Hair") and Stylespace\cite{Wu2020StyleSpaceAD} directions ("Blonde Hair", "Gender"). For car domain we used InterfaceGAN\cite{shen2020interpreting} ("Cube Shape", "Grass", "Colour Change") and Stylespace\cite{Wu2020StyleSpaceAD} ("Trees", "Headlights"). For each direction, we empirically choose several editing powers in such a way that $E$ produces non-artefacting edits. 

\section{Architecture details}\label{apend:arch}

\begin{figure}[htbp]
    \centering 

   \includegraphics[width=0.3\textwidth]{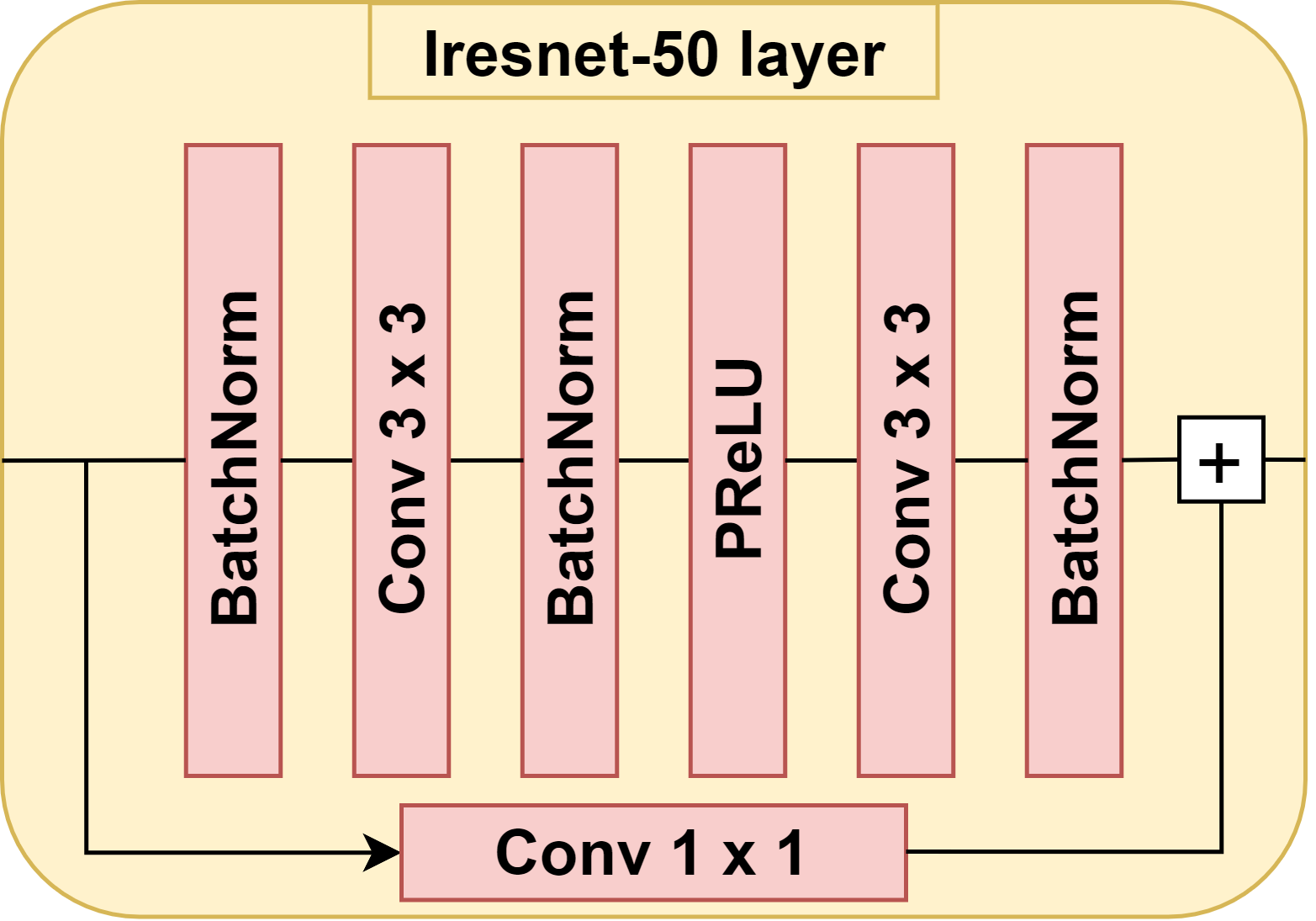}\\

    \caption{Scheme of the typical layer of Iresnet-50. $H$ and $I_{fus}$ also consist of such layers.}
    \label{app:layer}
\end{figure}

\begin{figure*}[htbp]
    \centering 
    \large
    \begin{tabular}{cc}

   \rotatebox{90}{~~~~~~~(a) Original }  &
   \includegraphics[width=\textwidth]{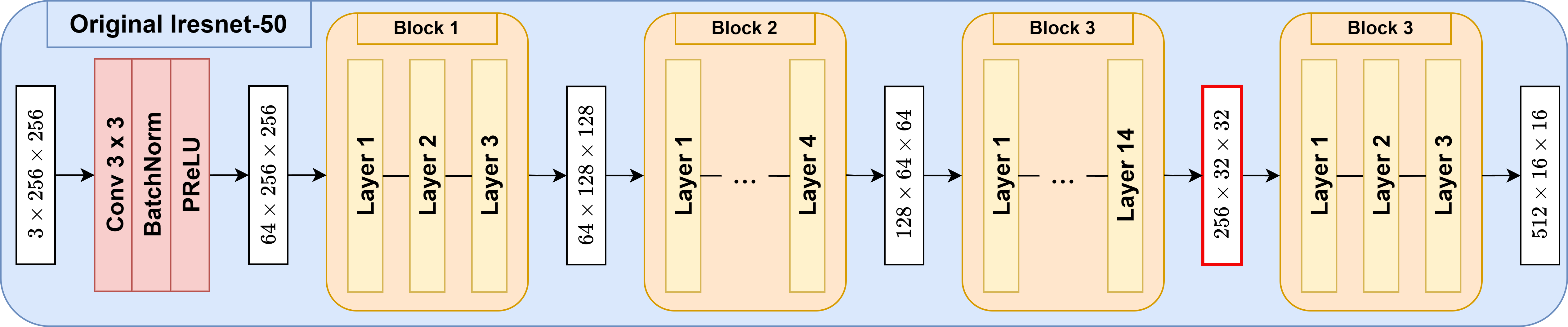}\\
   \label{fig:ablat_iris_orig}

   \rotatebox{90}{~~~~~~~~~~(b) Ours}  &
   \includegraphics[width=\textwidth]{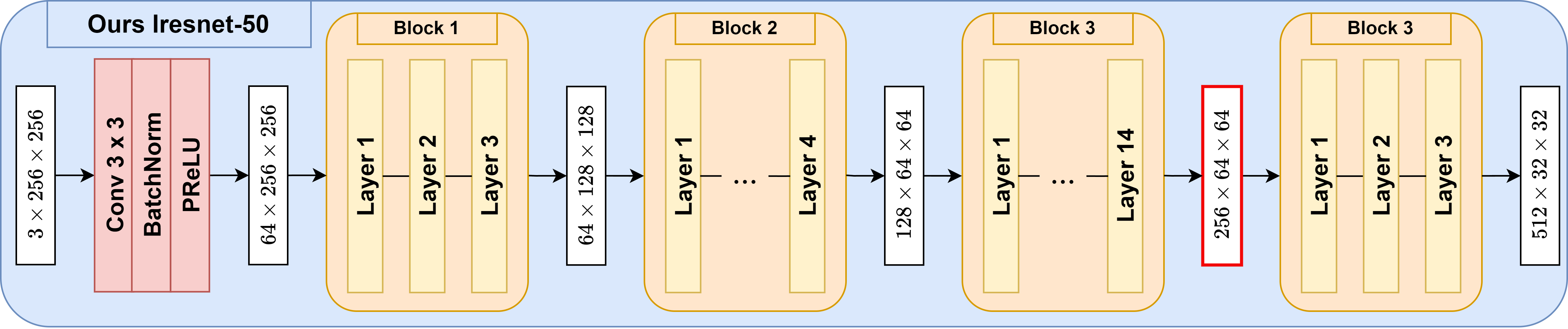}\\
   \end{tabular}

    \caption{Architecture of Iresnet-50 backbone. Red-framed output is the one that is then passed to Feature predictor to predict $F_{pred}$.}
    \label{app:iris}
\end{figure*}

Our architecture has 2 parts: Inverter $I$ and Feature Editor $H$. Inverter consists of Feature-Style-like encoder $I_{fse}$ and Fuser $I_{fus}$.  $I_{fse}$ has been slightly changed compared to original version. The original Iresnet-50 backbone consists of 4 blocks (Figure \ref{app:iris} (a)), where each block increases the number of channels and reduces the spatial resolution of the input tensor. Each block consists of several layers, whose typical architecture is shown in Figure \ref{app:layer}. As far as we increased $k$ from 5 to 9 (which increases spatial resolution of predicted tensor from $16 \times 16$ to $64 \times 64$) it is necessary to extract features from the backbone with corresponding to the new spatial resolution. However, in the original Iresnet-50 architecture, such a tensor can only be gathered after  block 2, which means that the original image is only passed through 3 + 4 = 7 layers, which is not enough to extract finer detail information. To fix this, we reduced stride in one of the layers in block 3, so that the resolution of the predicted tensor is changed as shown in Figure \ref{app:iris} (b), and the source image is processed through 21 layers.

It is important to note that in the car domain, the information of some editings consists only in high-dimensional features with a spatial resolution of $128 \times 128$. To take this into account, during $\Delta$ computation, we synthesize outputs of the 11-th generator layer $F_{w_E}, F_{w_E}' \in \mathcal{F}_{11}$ instead of $\mathcal{F}_9$. To transform such $\Delta$ to size of $ 512 \times 64 \times 64$, we first apply an additional trainable Irensnet-50 layer, which reduces the resolution, and only then pass processed $\Delta$ to $H$.

$I_{fus}$ and $H$ have the same architecture. They both consist of 6 Iresnet-50 layers (Figure \ref{app:layer}) with skip connections. During passing through $I_{fus}$ or $H$ spatial resolution of input tensor is not changed. When applying skip connections, we also use $1 \times 1$ convolution in case the number of feature map channels changes.

\begin{table}[t]
    \centering
    \caption{Additional editing results for StyleRes, SFE and checkpoint of SFE trained on a restricted set of editing directions $\mathcal{D}_{xmall}$ (see Ablation Study \ref{ablat} and Appendix \ref{app:edit}) on Celeba HQ. The technique used to calculate the editing metric is described in \ref{quant}. *However, since Celeba HQ does not have a rotation attribute, we used a different technique for this direction. We randomly divided Celeba HQ into 2 equal parts, applied rotation to one of them and calculated the FID between them to evaluate the realism of the edited images.}
    
    \label{app:reb_edit}
    \Large
    \resizebox{1.0\linewidth}{!}{
    \begin{tabular}{lccccc}
        \toprule
        
        Model &  Glasses(+)  & Rotation*(-) & Rotation*(+) & Bangs (+) & Beard(+)\\
        \midrule
         StyleRes & \textbf{73.089}  & 26.004 & 27.492 & 45.497 & 77.084\\
          $\mathcal{D}_{small}$ & 74.855  &  \underline{25.429} & \underline{26.371} & \underline{40.006} & \textbf{76.168}\\
          SFE & \underline{73.098}  &  \textbf{23.541} & \textbf{24.084} & \textbf{39.319} & \underline{77.529}\\
          \bottomrule
    \end{tabular}
}
\end{table}

\section{Masking} \label{app:mask_sec}
To edit images, StyleFeatureEditor uses editing information from the additional encoder $E$, which allows Inverter to focus only on reconstruction features. However, this is also a disadvantage of our method: if $w_E$ would have artefacts during editing, Feature Editor will mostly inherit these artefacts. Therefore, it is important to choose $E$ carefully. Unfortunately, there is a more general problem: some directions may not only change the attribute to which they refer, but also influence others.

Typically, 2 types of artefacts appear. First, while editing one attribute, another face attribute may be changed.  For example, when adding glasses with a higher editing power, the mouth starts to open. The second type is that because $E$ is only a w-latent encoder, it cannot reconstruct background well and make it smooth, so during editing, such background could also be affected (for example directions of bob cut and bowl cut hairstyles).

\begin{figure*}[htbp]
    \centering 
    \large
    \begin{tabular}{cc}
   \rotatebox{90}{\hspace{3.5mm} Inversion}  &
   \includegraphics[width=\linewidth]{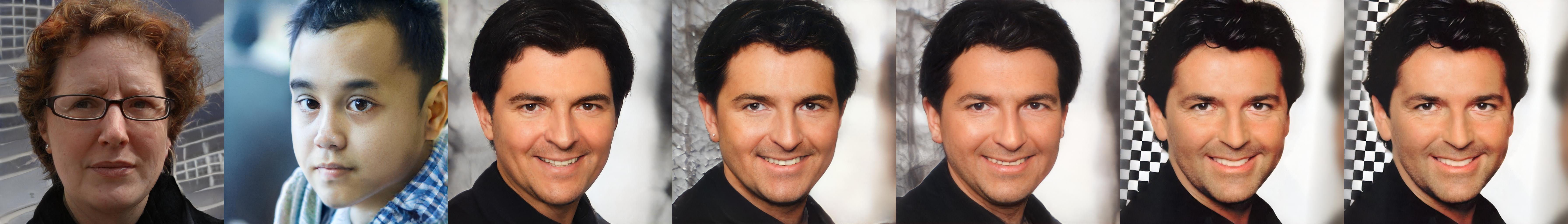}\vspace{-0.1cm}\\ 

   \rotatebox{90}{\hspace{5mm} Bobcut}  &
   \includegraphics[width= \linewidth]{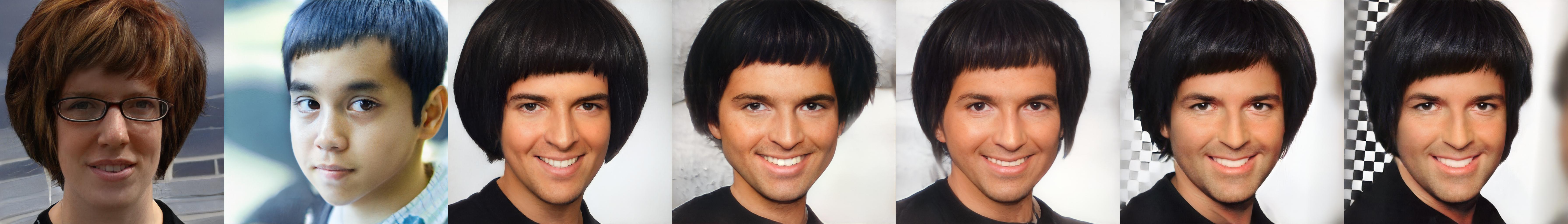}\vspace{0.3cm}\\
    
    \rotatebox{90}{\hspace{3.5mm} Inversion}  &
   \includegraphics[width= \linewidth]{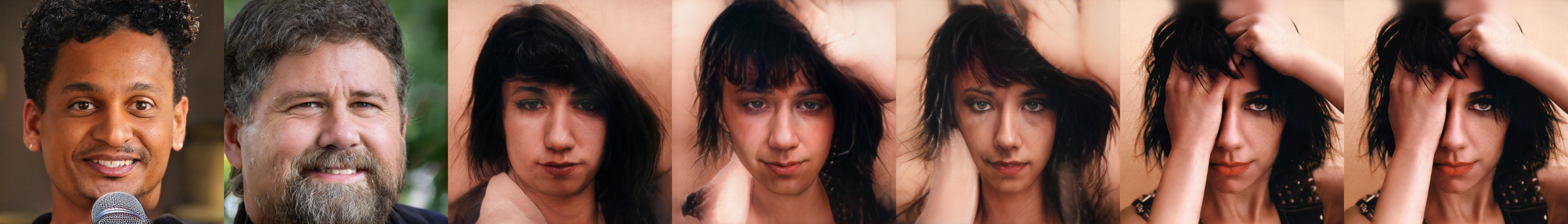}\vspace{-0.1cm}\\
   
   \rotatebox{90}{\hspace{4.5mm} Glasses}  &
   \includegraphics[width=\linewidth]{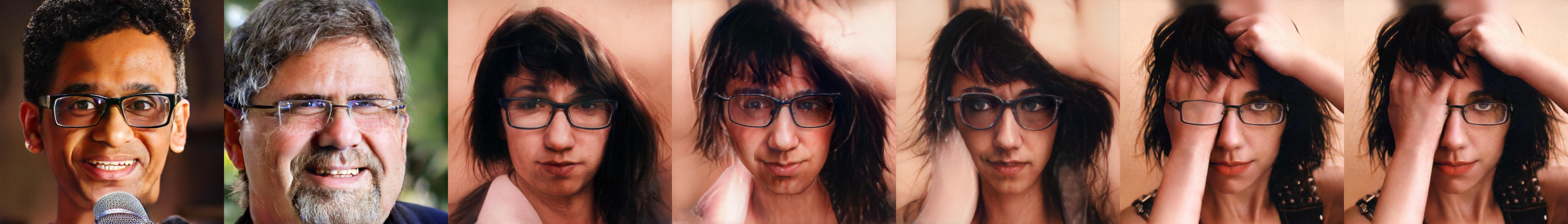}\vspace{0.3cm}\\

    \rotatebox{90}{\hspace{3.5mm} Inversion}  &
   \includegraphics[width= \linewidth]{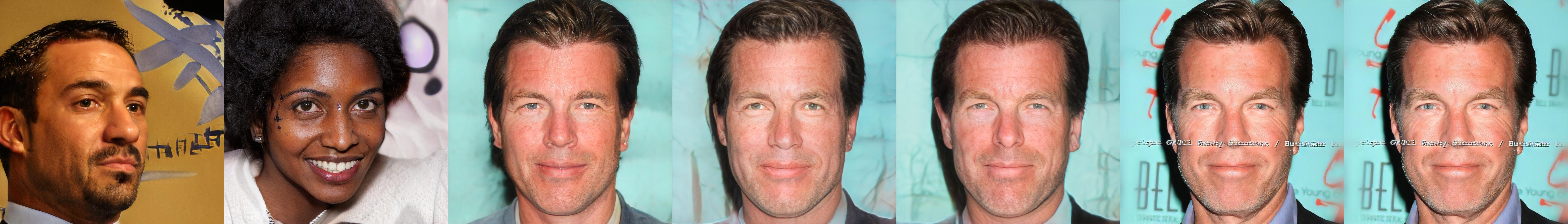}\vspace{-0.1cm}\\
   
   \rotatebox{90}{\hspace{3.5mm} Rotation}  &
   \includegraphics[width=\linewidth]{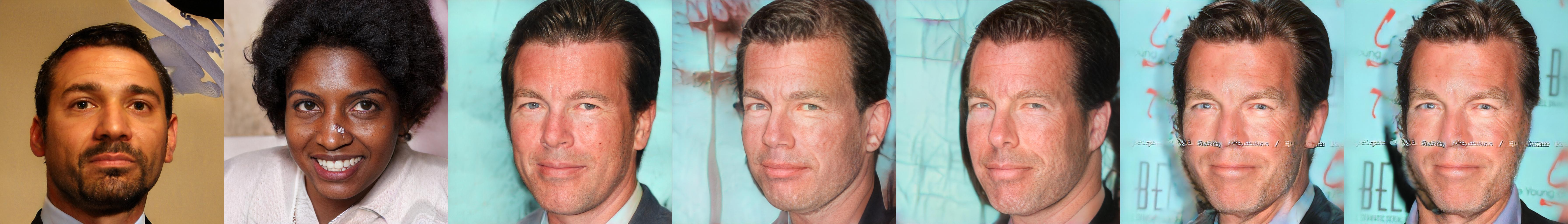}\\

    \end{tabular}
    \begin{tabular*}{1.05\linewidth}{@{\extracolsep{\fill}}cccccccc}

     & \hspace{4mm} Synthetic 1 & \hspace{0mm} Synthetic 2 & \hspace{10mm} e4e & \hspace{13mm} pSp & \hspace{5mm} StyleTrans & \hspace{7mm} Ours & \hspace{3mm} Ours masked \hspace{2mm} \\
    \end{tabular*}
    \caption{Examples of artefacts created by inaccurate editing directions. The first two columns represent synthetic images synthesized from $w$ (Inversion) or its edited version $w'$ (Corresponding Editing Direction), where $w$ is obtained by randomly sampling $z$ and passing it through the StyleGAN Mapping Network. Other columns represent inversion of real images (they are not represented here, but they are visually indistinguishable from the inversion of Ours method) by different encoders and its edited versions. During Bobcut editing, the background starts to disappear (even for synthetic images); during Glasses editing, the mouth starts to open. The masking technique (last column, for more details see Section \ref{app:mask_sec}) allows our method to avoid artefacts that appear during editing within the face zone (Bobcut, Glasses), but does not allow editing correctly while regions outside the face zone should be edited (Rotation).}
    \label{app:mask}
\end{figure*}

Our method is able to fix the second type of such artefacts. To do this, we propose to use an additional pre-trained model $M$, which is able to predict the face mask of the source image. The mask is scaled to a resolution of $64 \times 64$ and applied to $\Delta$ so that all features outside the face zone become zeros. As $\Delta$ preserves positional information, this means that the part of the image outside the face zone  is not edited. The results of this approach are shown in Figure \ref{app:mask}.

However, such a simple technique could lead to artefacts in cases where editing should be applied outside the face zone,  such as pose rotation or afro hairstyle. Therefore, we left this technique as an optional feature. 

\section{Editings generaization}\label{app:edit}
\begin{figure*}[htbp]

 \centering
    \vspace{-0.25cm}
    \small
    \renewcommand{\arraystretch}{0}
    \begin{tabular}{cc}
    \rotatebox{90}{\hspace{0.4mm} StyleRes}  & \includegraphics[width= 1\linewidth]{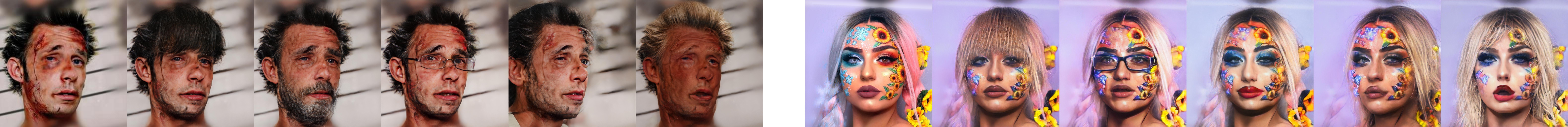}\\

    \rotatebox{90}{\hspace{1.5mm} $\mathcal{D}_{small}$}  & \includegraphics[width= 1\linewidth]{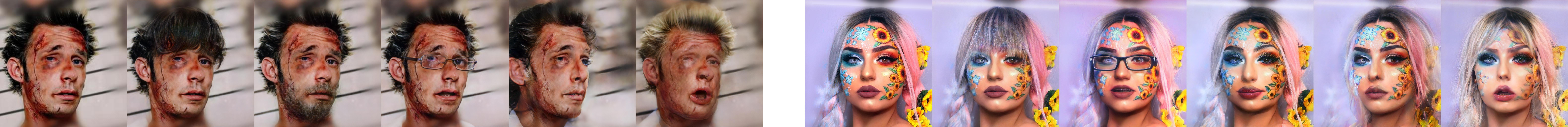}\\

   \rotatebox{90}{\hspace{3.2mm} SFE}  & \includegraphics[width= 1\linewidth]{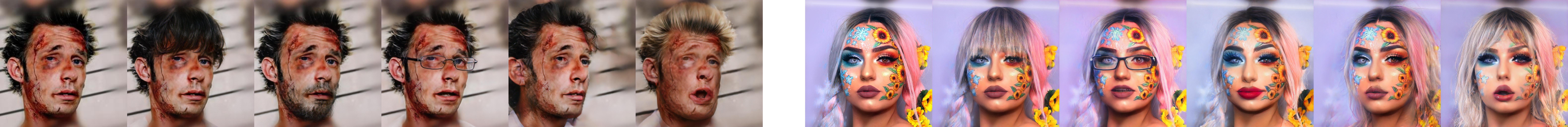}\\

    \end{tabular}

    \begin{tabular*}{1.0\linewidth}{@{\extracolsep{\fill}}ccccccccllll}
     \hspace{7mm} Original  & \hspace{1mm} Bangs & \hspace{4mm} Beard &  \hspace{2.5mm} Glasses & \hspace{1.5mm} Rotation & \hspace{1.5mm} Trump \hspace{5.5mm}& \hspace{9mm} Orig & \hspace{3.5mm} Bangs & \hspace{2mm} Glasses & \hspace{0.5mm} Make-Up & \hspace{0mm} Rotation  \hspace{0mm}&  \footnotesize{Taylor-Swift\hspace{0mm}}\\
    \end{tabular*}

    \captionof{figure}{Additional editing results for SFE, StyleRes and SFE trained on stricted set of edits $\mathcal{D}_{small}$ (see question 1). Better zoom-in.}

    \label{app:add_compare}

\end{figure*}

In this section we provide additional results from the Ablation Study checkpoint $\mathcal{D}_{small}$. This checkpoint was trained on a restricted set of editing directions $\mathcal{D}_{small}$ (see Ablation Study \ref{ablat}). In Figure \ref{app:add_compare} we compare this checkpoint with StyleRes and our main model (SFE) on directions not presented in $\mathcal{D}_{small}$. Both of our models outperform StyleRes, preserving more image detail and providing comparable editing, while the restricted and unrestricted checkpoints are only slightly different. This proves that our method is generalisable to any direction, even those not represented in the training set. The numerical results in Tab. \ref{app:reb_edit} also confirm this.

\section{Additional results}
\begin{figure*}[t]
    \centering 
    \large
    \renewcommand{\arraystretch}{0}
    \begin{tabular}{cc}
    \rotatebox{90}{ \hspace{5.5mm} Original}  & \includegraphics[width= 1\linewidth]{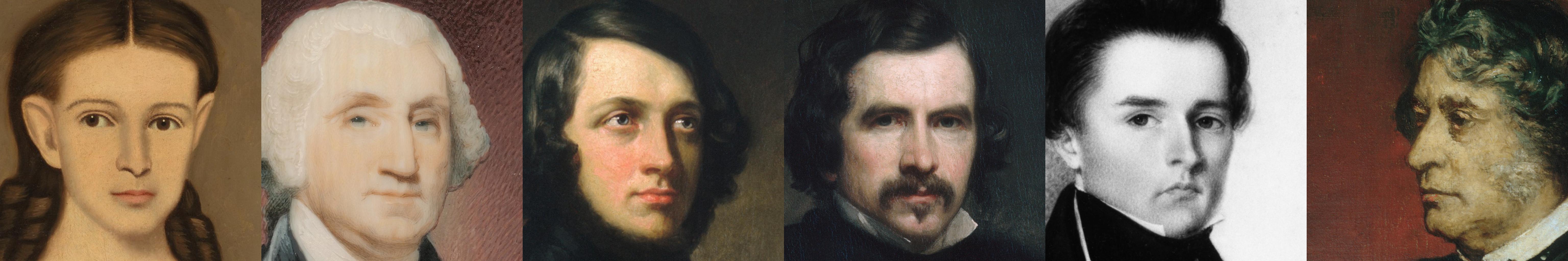}\\

    \rotatebox{90}{\hspace{5.5mm} StyleRes}  & \includegraphics[width= 1\linewidth]{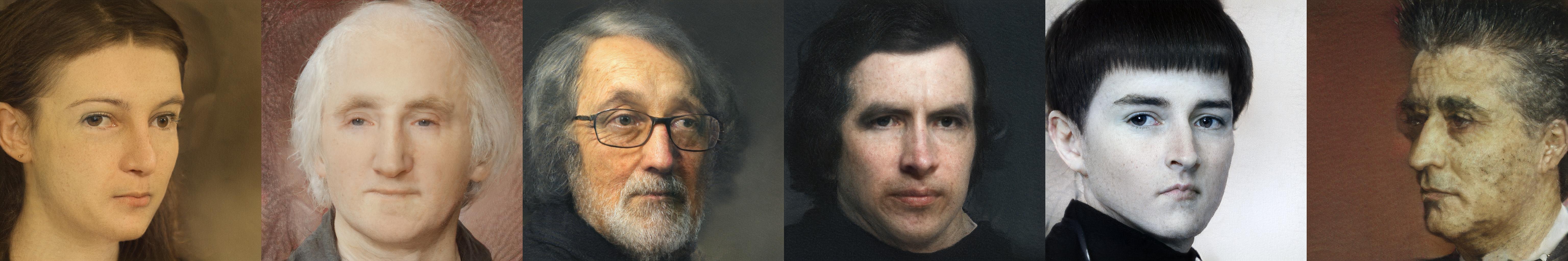}\\
    
    \rotatebox{90}{\hspace{10mm} SFE}  & \includegraphics[width= 1\linewidth]{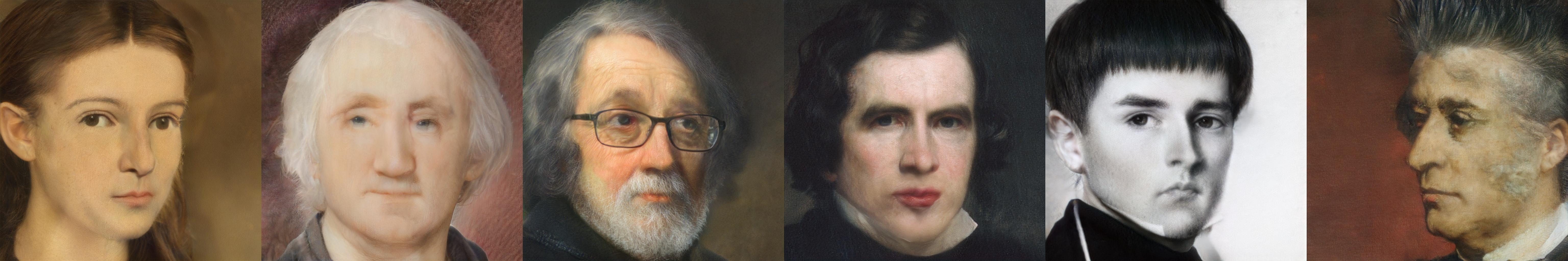}\\

   \end{tabular}
    
    \begin{tabular}{cccccc}
    \hspace{10mm}
     Rotation(+) & \hspace{7mm} 
     Rotation(-) & \hspace{11mm} 
     Age(+) & \hspace{12mm} 
     Beard(-) & \hspace{10mm} 
     Bowlcut & \hspace{13mm} 
     Mohawk
    \end{tabular}
    \caption{Results for SFE and StyleRes on MetaFaces. SFE preserves the original image style, while StyleRes makes it more photorealistic (see 5th and 6th columns).}
    \label{fig:reb_meta}
\end{figure*}

\begin{figure*}[!htbp]
    \centering 
    \begin{tabular*}{0.9\linewidth}{@{\extracolsep{\fill}}ccccccccc}
     
     \hspace{5mm}Input  & \hspace{2mm} Inversion & \hspace{2mm} Age(+) & \hspace{2mm} Smile(+) & Smile(-) & 
 \hspace{1mm} Glasses \hspace{3mm} & Bobcut & Blond Hair & Dark Hair\\
    \end{tabular*}

    \begin{tabular}{c}
     \includegraphics[width=0.9\linewidth]{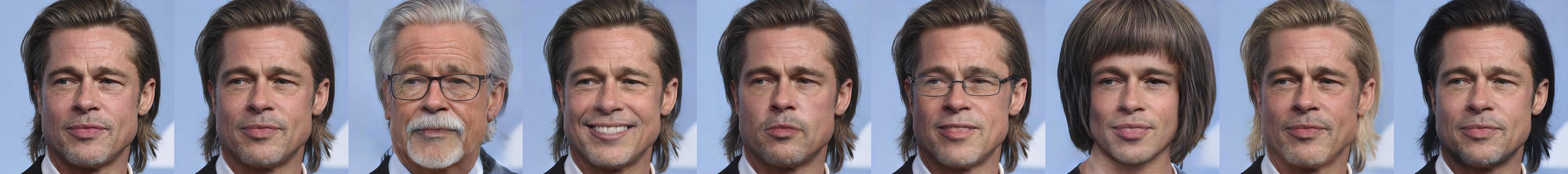}\\
     \includegraphics[width=0.9\linewidth]{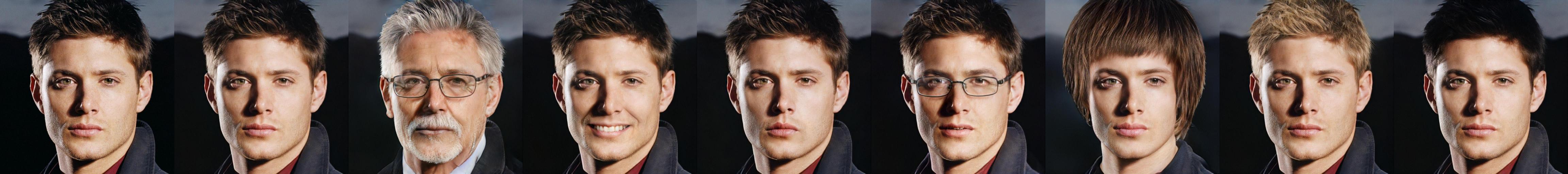}\\
     \includegraphics[width=0.9\linewidth]{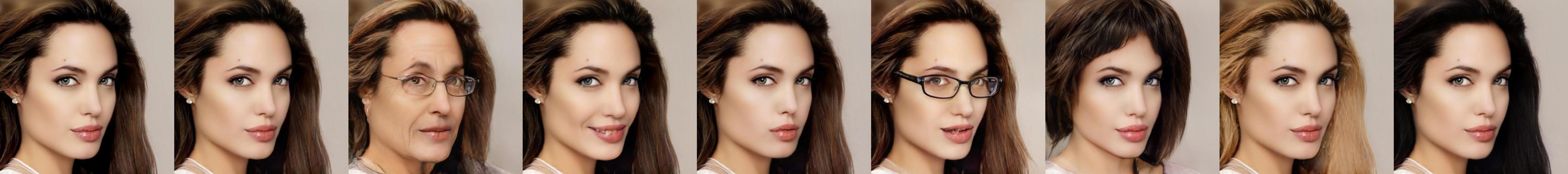}\\
     \includegraphics[width=0.9\linewidth]{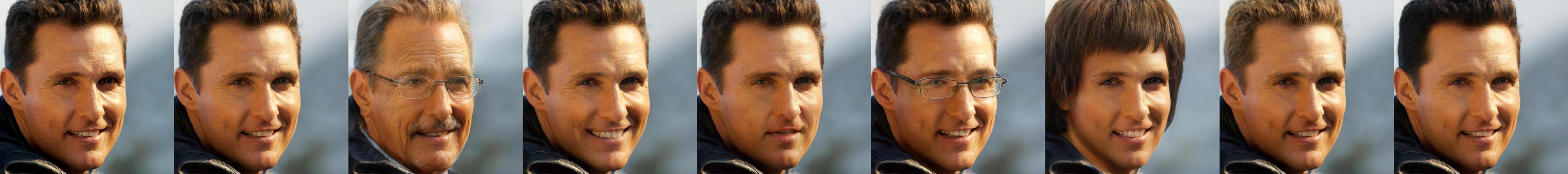}\\
     \includegraphics[width=0.9\linewidth]{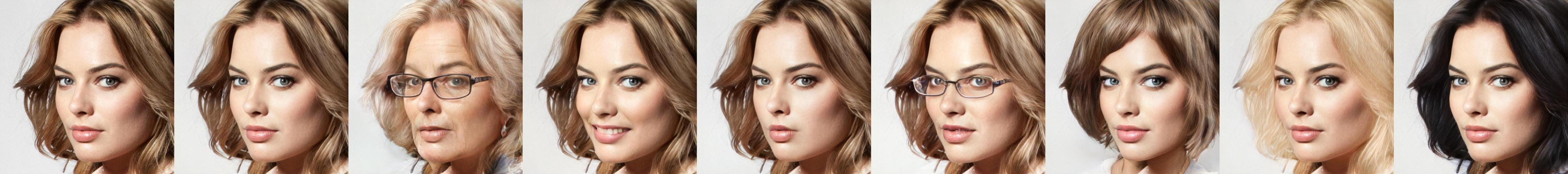}\\
     \includegraphics[width=0.9\linewidth]{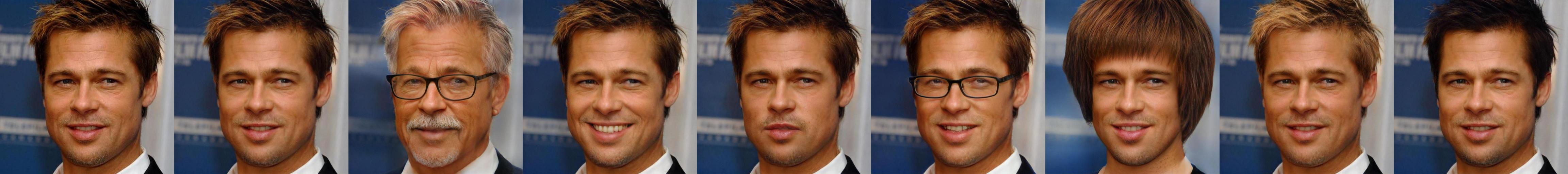}\\
     \includegraphics[width=0.9\linewidth]{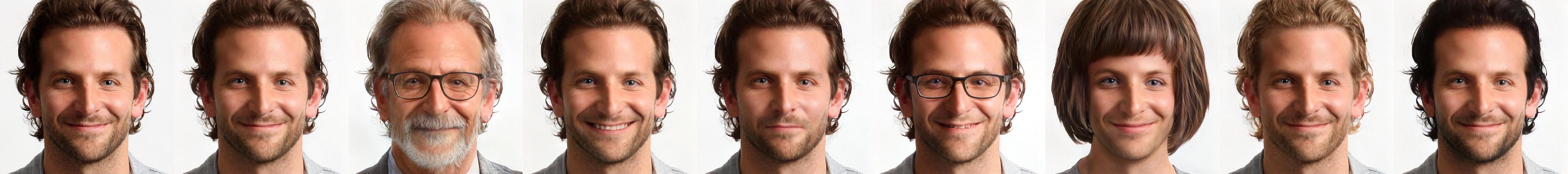}\\
     \includegraphics[width=0.9\linewidth]{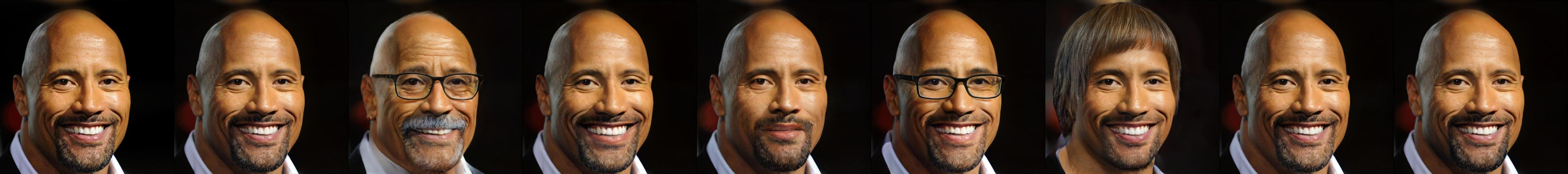}\\
     \includegraphics[width=0.9\linewidth]{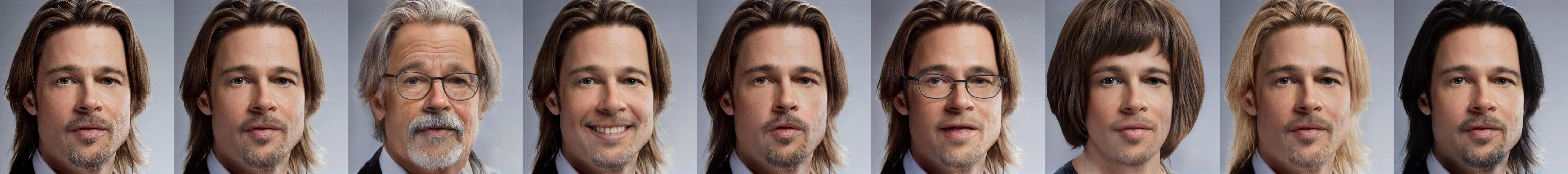}\\
     \includegraphics[width=0.9\linewidth]{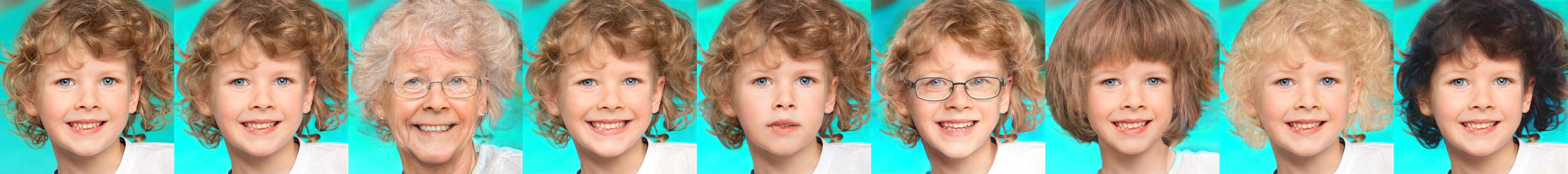}\\
     \includegraphics[width=0.9\linewidth]{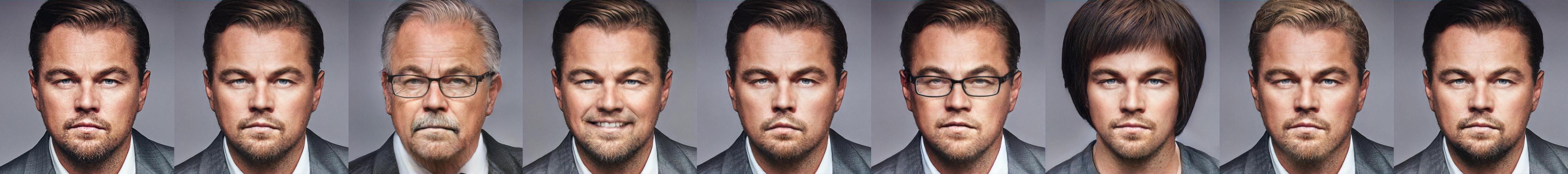}\\

    \end{tabular}
    \begin{tabular*}{0.9\linewidth}{@{\extracolsep{\fill}}ccccccccc}
     
     \hspace{5mm}Input  & \hspace{2mm} Inversion & \hspace{2mm} Age(+) & \hspace{2mm} Smile(+) & Smile(-) & 
 \hspace{1mm} Glasses \hspace{3mm} & Bobcut & Blond Hair & Dark Hair\\
    \end{tabular*}
    \caption{Additional visual example of StyleFeatureEditor in face domain.}
    \label{app:edit_our_face}
\end{figure*}

\begin{figure*}[!htbp]
    \centering 
    \begin{tabular}{cc}
    \rotatebox{90}{\hspace{5mm} Afro}  & \includegraphics[width=0.8\linewidth]{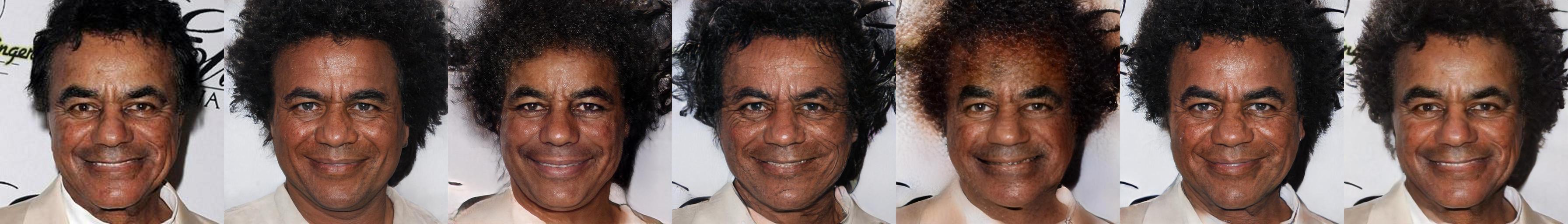}\\

    \rotatebox{90}{\hspace{3.5mm} Young}  & \includegraphics[width= 0.8\linewidth]{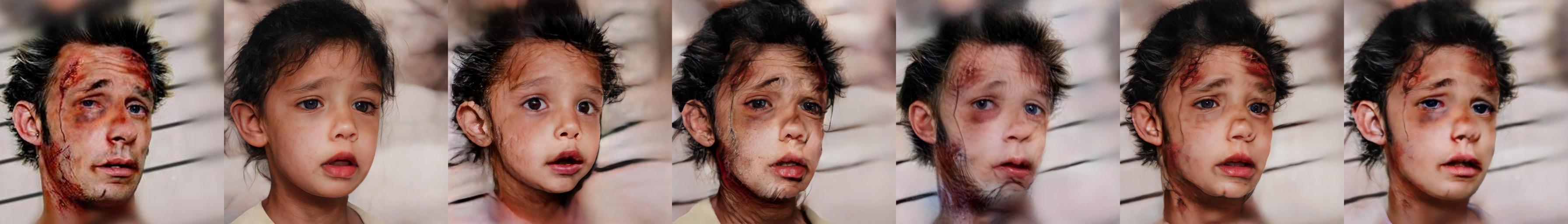}\\

   \rotatebox{90}{\hspace{4.5mm} Angry}  & \includegraphics[width= 0.8\linewidth]{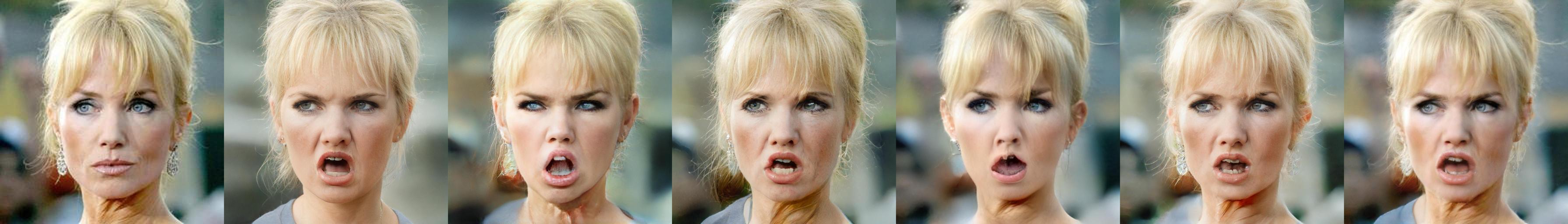}\\
   
   \rotatebox{90}{ Eye openness}  & 
   \includegraphics[width=0.8 \linewidth]{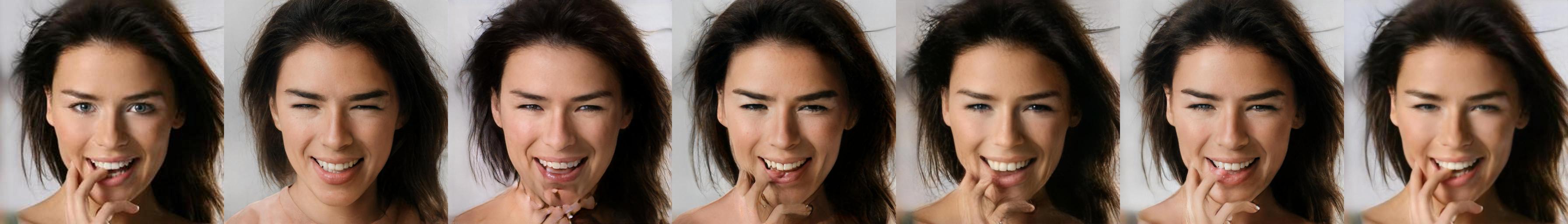}\\ 

   \rotatebox{90}{\hspace{5mm} Gender}  & 
   \includegraphics[width=0.8\linewidth]{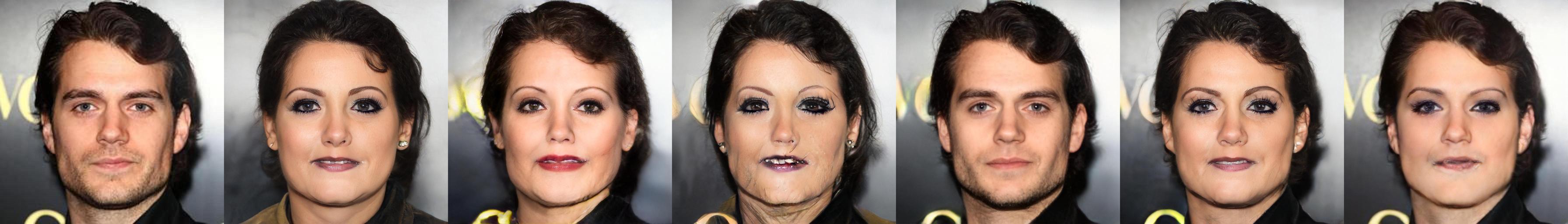}\\

   \rotatebox{90}{\hspace{5mm} Goatee}  & 
   \includegraphics[width=0.8\linewidth]{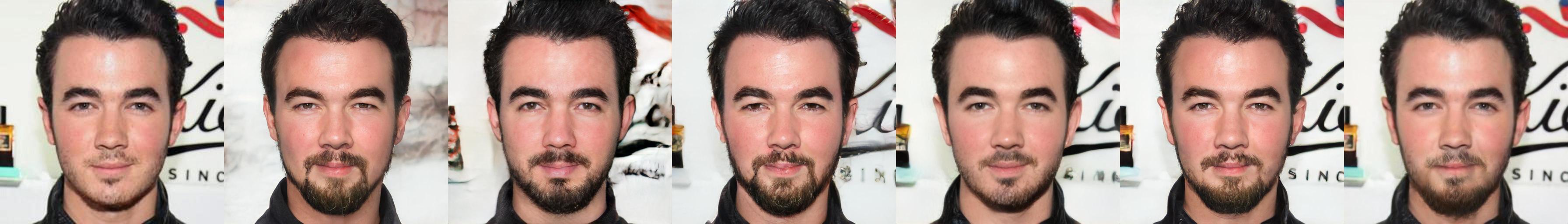}\\

   \rotatebox{90}{\hspace{2mm} Inversion}  & 
   \includegraphics[width=0.8\linewidth]{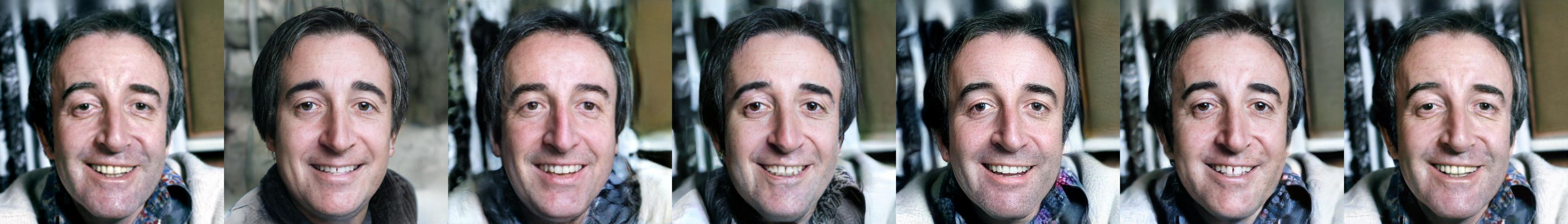}\\

   \rotatebox{90}{\hspace{3mm} Make-up}  & 
   \includegraphics[width=0.8\linewidth]{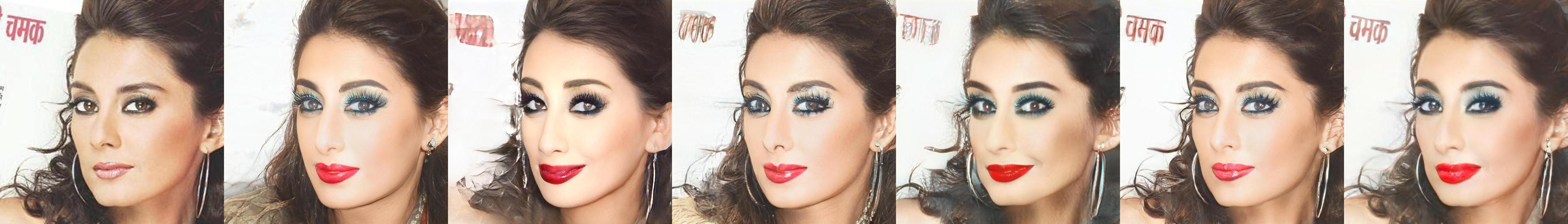}\\

   \rotatebox{90}{\hspace{1mm}  Purple Hair}  & 
   \includegraphics[width=0.8\linewidth]{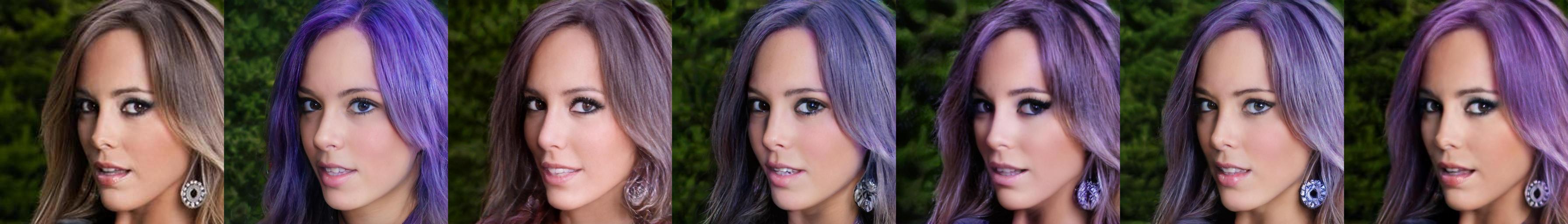}\\

   \rotatebox{90}{\hspace{2mm} Rotation}  & 
   \includegraphics[width=0.8\linewidth]{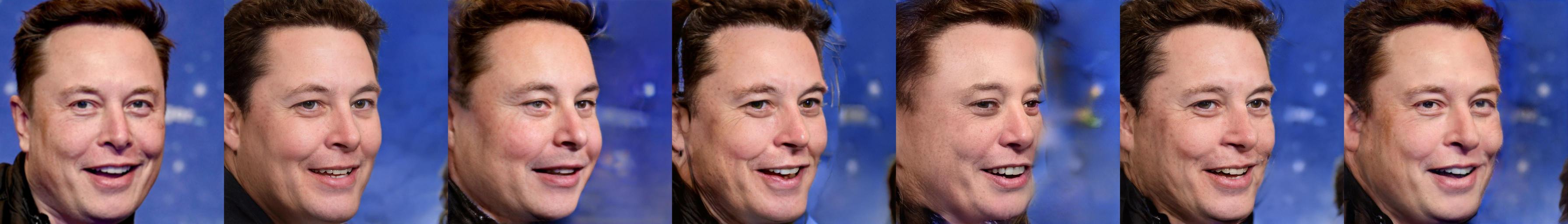}\\

    \end{tabular}
    \begin{tabular*}{0.8\linewidth}{@{\extracolsep{\fill}}cccccrrr}
     & \hspace{6mm} Input &  \hspace{10mm} e4e &  \hspace{6mm}Hyperinverter \hspace{2mm} & HFGI \hspace{3mm} & \hspace{10mm}FS & \hspace{10mm}StyleRes & \hspace{4mm}SFE (ours)
    \end{tabular*}
    \caption{Additional visual comparison of FaetureStyleEditor with previous approaches in face domain.}
    \label{app:edit_comp_face}
\end{figure*}

\begin{figure*}[htbp]
    \centering 
    \begin{tabular}{cc}

   \rotatebox{90}{ Inversion 1}  &
   \includegraphics[width= 0.9\linewidth]{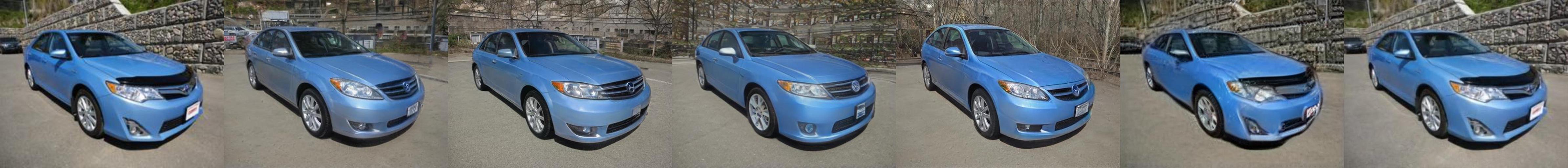}\\ 

   \rotatebox{90}{ Inversion 2}  &
   \includegraphics[width= 0.9\linewidth]{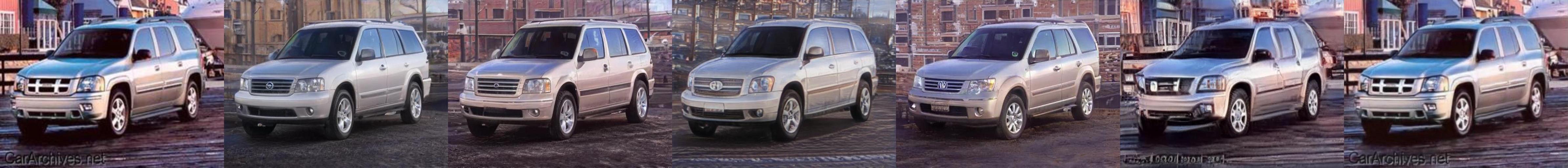}\\

    \rotatebox{90}{\hspace{3.5mm} Cube 1}  &
   \includegraphics[width= 0.9\linewidth]{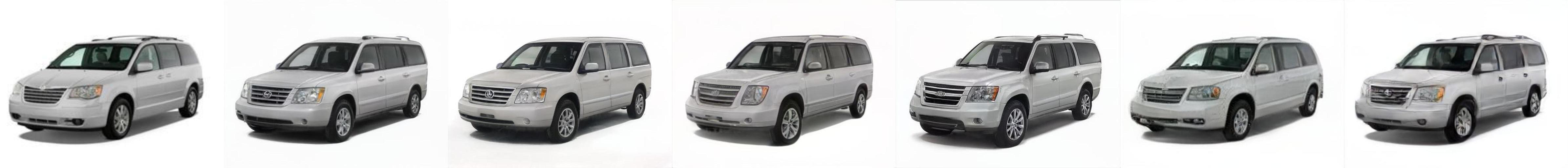}\\

   \rotatebox{90}{\hspace{3.5mm} Cube 2}  &
   \includegraphics[width= 0.9\linewidth]{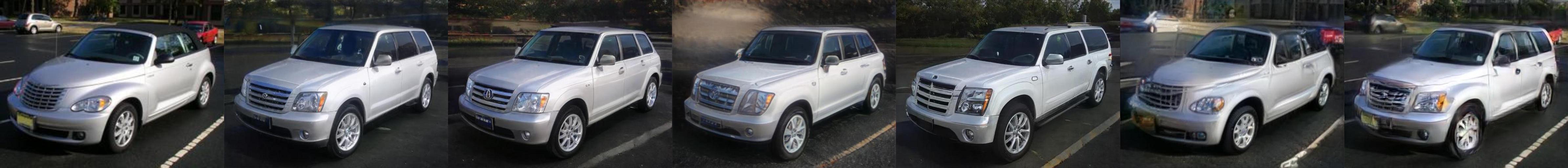}\\

   \rotatebox{90}{\hspace{2mm} Colour 1}  &
   \includegraphics[width= 0.9\linewidth]{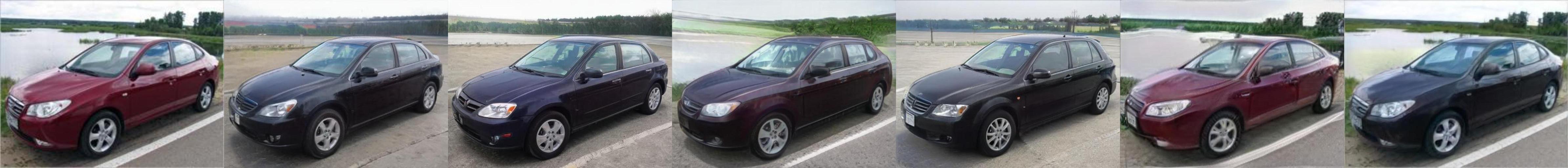}\\

   \rotatebox{90}{\hspace{2mm} Colour 2}  &
   \includegraphics[width= 0.9\linewidth]{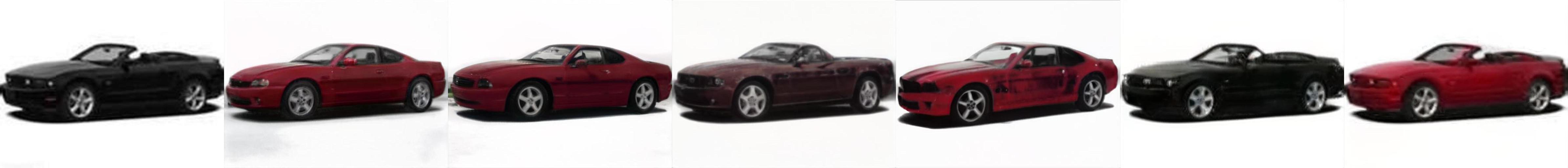}\\

   \rotatebox{90}{\hspace{3mm} Grass 1}  &
   \includegraphics[width= 0.9\linewidth]{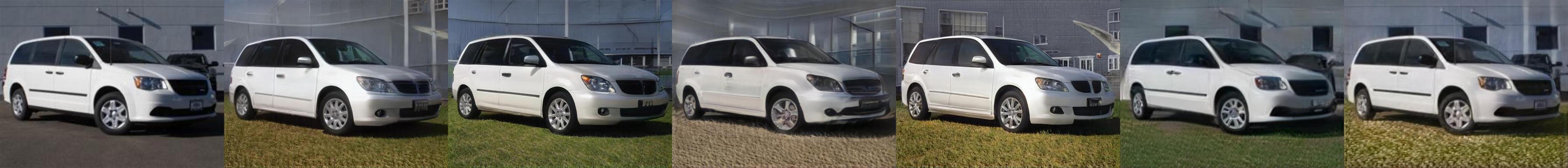}\\

   \rotatebox{90}{\hspace{3mm} Grass 2}  &
   \includegraphics[width= 0.9\linewidth]{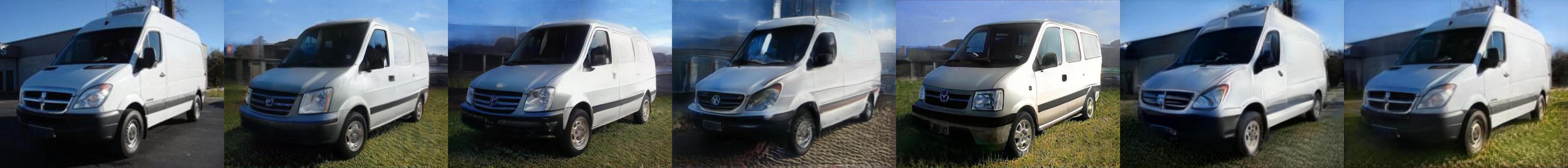}\\

   \rotatebox{90}{\hspace{2.5mm} Lights 1}  &
   \includegraphics[width= 0.9\linewidth]{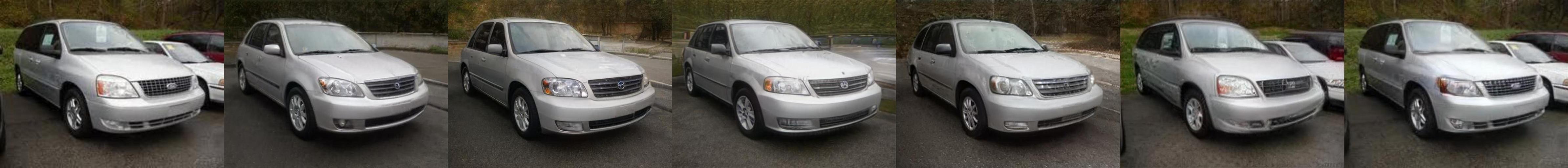}\\

   \rotatebox{90}{\hspace{1.5mm} Lights 2}  &
   \includegraphics[width= 0.9\linewidth]{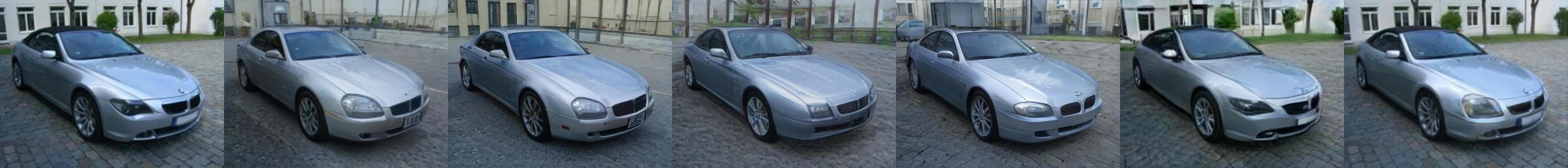}\\

    \end{tabular}
    \begin{tabular*}{0.9\linewidth}{@{\extracolsep{\fill}}cccccccc}

    & \hspace{4mm} Input &  \hspace{12mm} e4e &  \hspace{8mm} ReStyle & \hspace{2mm} StyleTrans \hspace{1mm} &  HyperStyle & \hspace{5mm} FS & \hspace{6mm} SFE (ours) \\
    \end{tabular*}
    \caption{Additional visual comparison of StyleFaetureEditor with previous approaches in car domain.}
    \label{app:edit_car}
\end{figure*}

In this section we provide additional visual examples of the StyleFeatureEditor. In Figure \ref{fig:reb_meta} we compare our method with StyleRes on out-of-domain MetaFaces dataset. Our method is able to preserve the original image style, while StyleRes makes it more realistic.
In Figure \ref{app:edit_our_face} we show the work of our method in the face domain for several additional editing directions. In Figure \ref{app:edit_comp_face} we present an additional comparison between StyleFeatureEditor and previous approaches in the face domain, and in Figure \ref{app:edit_car} we present more results for the car domain.

\end{document}